\documentclass[11pt, logo, copyright]{nvidiatechreport}
\usepackage{graphicx}
\usepackage{wrapfig}
\usepackage{subcaption,ragged2e}
\usepackage[round]{natbib}
\usepackage{booktabs}
\usepackage{tabularx}
\usepackage{listings}
\usepackage{xcolor}
\usepackage{upquote}
\usepackage{tikz}
\usetikzlibrary{arrows.meta,calc,fit,positioning,shapes.geometric}
\usepackage{amsmath}
\usepackage{placeins}
\usepackage{xurl}
\usepackage{hyperref}
\usepackage{xspace}
\hypersetup{colorlinks=true, allcolors=blue}
\renewcommand{\absfont}{\normalfont\normalsize}


\definecolor{ddgreen}{HTML}{76B900}
\definecolor{dddeepgreen}{HTML}{2E6F2E}
\definecolor{ddblue}{HTML}{1F77B4}
\definecolor{ddteal}{HTML}{00897B}
\definecolor{ddorange}{HTML}{D97A00}
\definecolor{ddpurple}{HTML}{6F4BB2}
\definecolor{ddred}{HTML}{B24A3A}
\definecolor{ddink}{HTML}{1F2933}
\definecolor{ddmuted}{HTML}{5D6778}
\definecolor{ddline}{HTML}{CAD4C5}
\definecolor{ddpalegreen}{HTML}{EFF8E8}
\definecolor{ddpaleblue}{HTML}{EEF6FC}
\definecolor{ddpaleteal}{HTML}{EAF7F5}
\definecolor{ddpalewarm}{HTML}{FFF5E6}
\definecolor{ddcodebg}{HTML}{F8FAF6}
\definecolor{ddcodeframe}{HTML}{B6C8AD}
\definecolor{ddcodekw}{HTML}{1F5E99}
\definecolor{ddcodestr}{HTML}{2E7D32}
\definecolor{ddcodecomment}{HTML}{6B7280}
\definecolor{ddcodeemph}{HTML}{7A3E9D}

\lstdefinelanguage{DDSh}{
  sensitive=true,
  morekeywords={pip,export,data-designer,config,list},
  morecomment=[l]{\#},
  morestring=[b]",
  morestring=[b]'
}

\lstdefinelanguage{DDYaml}{
  sensitive=false,
  alsoletter={_,-},
  morekeywords={columns,processors,name,column_type,processor_type,sampler,sampler_type,category,params,values,llm-text,model_alias,prompt,schema_transform,template,messages,role,content,user,assistant},
  morecomment=[l]{\#},
  morestring=[b]",
  morestring=[b]'
}

\lstdefinestyle{ddlisting}{
  basicstyle=\ttfamily\footnotesize,
  backgroundcolor=\color{ddcodebg},
  breaklines=true,
  columns=fullflexible,
  frame=single,
  framerule=0.45pt,
  rulecolor=\color{ddcodeframe},
  keywordstyle=\color{ddcodekw}\bfseries,
  stringstyle=\color{ddcodestr},
  commentstyle=\color{ddcodecomment}\itshape,
  emph={DataDesigner,DataDesignerConfigBuilder,SamplerColumnConfig,LLMTextColumnConfig,LLMCodeColumnConfig,LLMStructuredColumnConfig,LLMJudgeColumnConfig,ValidationColumnConfig,ToolConfig,DropColumnsProcessorConfig,SchemaTransformProcessorConfig,SamplerType,CategorySamplerParams,CodeValidatorParams,CodeLang,TraceType,BaseModel,Field},
  emphstyle=\color{ddcodeemph}\bfseries,
  keepspaces=true,
  showstringspaces=false,
  tabsize=2,
  upquote=true,
  xleftmargin=0.5em,
  xrightmargin=0.5em,
  framexleftmargin=0.5em,
  framexrightmargin=0.5em,
  aboveskip=0.8\baselineskip,
  belowskip=0.8\baselineskip,
  captionpos=b
}
\tikzset{
  ddfig/.style={
    font=\small\sffamily,
    >=Latex,
    line width=0.55pt
  },
  ddtitle/.style={
    font=\bfseries\sffamily,
    text=ddink,
    align=center
  },
  ddbox/.style={
    draw=ddline,
    rounded corners=2pt,
    fill=white,
    align=center,
    inner sep=5pt,
    text=ddink
  },
  ddgreenbox/.style={ddbox, draw=ddgreen!70!black, fill=ddpalegreen},
  ddbluebox/.style={ddbox, draw=ddblue!65!black, fill=ddpaleblue},
  ddtealbox/.style={ddbox, draw=ddteal!65!black, fill=ddpaleteal},
  ddwarmbox/.style={ddbox, draw=ddorange!70!black, fill=ddpalewarm},
  ddpurplebox/.style={ddbox, draw=ddpurple!65!black, fill=ddpurple!7},
  ddredbox/.style={ddbox, draw=ddred!70!black, fill=ddred!7},
  ddarrow/.style={->, draw=ddmuted, line width=0.7pt},
  ddstrongarrow/.style={->, draw=dddeepgreen, line width=1.0pt},
  ddnote/.style={font=\scriptsize\sffamily, text=ddmuted, align=center}
}

\newcommand{\githubicon}{%
  \raisebox{-0.18em}{\includegraphics[height=1.1em]{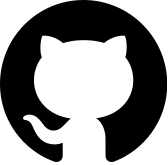}}%
}

\newcommand{\library}{NeMo Data Designer\xspace}
\newcommand{\ndd}{NDD\xspace}

\title{\library: An Extensible Framework for Multimodal Synthetic Data Generation}
\author{Johnny Greco, Nabin Mulepati, Andre Manoel, Eric Tramel, Kirit Thadaka,
Mike Knepper, Dhruv Nathawani, Dane Corneil, Yev Meyer, Alex Watson, Maarten Van Segbroeck}
\newcommand{\authorgap}{\hspace{2.4em}}
\newcommand{\reportcontributors}{%
\setlength{\baselineskip}{1.55em}%
\centering
\textbf{Johnny Greco}\authorgap\textbf{Nabin Mulepati}\authorgap\textbf{Andre Manoel}\\
\textbf{Eric Tramel}\authorgap\textbf{Kirit Thadaka}\authorgap\textbf{Mike Knepper}\authorgap\textbf{Dhruv Nathawani}\\
\textbf{Dane Corneil}\authorgap\textbf{Yev Meyer}\authorgap\textbf{Alex Watson}\authorgap\textbf{Maarten Van Segbroeck}}
\date{}

\makeatletter
\renewcommand{\maketitle}{%
  \bgroup
  \setlength{\parindent}{0pt}%
  \begin{center}
    {\titlefont
      NeMo Data Designer: An Extensible Framework\par
      for Multimodal Synthetic Data Generation\par}
    \vskip1.1em
    {\normalfont\bfseries\fontsize{10.5}{15}\selectfont
      \reportcontributors\par}
    \vskip0.75em
    {\normalfont\bfseries\fontsize{10}{12}\selectfont
      \href{https://github.com/NVIDIA-NeMo/DataDesigner}{%
        \githubicon\hspace{0.45em}%
        \textcolor{dddeepgreen}{\texttt{NVIDIA-NeMo/DataDesigner}}}\par}
  \end{center}
  \vskip1.4em
  \egroup
  {\abscontent}%
  \thispagestyle{firststyle}%
}
\makeatother

\begin{document}

\begin{abstract}
We present \textbf{\library} (\ndd), an open-source, general-purpose framework for multimodal synthetic data generation (SDG). Designed to be intuitive to use, \ndd provides a declarative configuration format in which human and/or agent users define each dataset column, with column types spanning text, code, structured outputs, images, embeddings, and statistical samplers that are explicitly configured to steer dataset diversity. Additional column types and functionality can be introduced using the framework's flexible plugin system. \ndd's configuration is an inspectable artifact, supporting workflow sharing and reproducibility. SDG is an inherently iterative process. \ndd therefore builds a preview-and-revision loop into its core workflow, allowing users to generate and inspect a small number of records, refine the specification, and rerun generation at full scale. At runtime, \ndd resolves dependencies, schedules calls to user-provided model endpoints, and retries failed requests. We describe \ndd's architecture and programming model and present case studies spanning structured, agentic, multimodal, and domain-specialized tasks, including datasets used in Nemotron model development and in production enterprise deployments.
\end{abstract}

\maketitle

\section{Introduction}
\label{sec:introduction}

Synthetic data generation (SDG) has long played an important role in science and technology. From simulating virtual neutron histories in Monte Carlo experiments~\citep{richtmyer1947neutron}, to boosting minority classes in imbalanced datasets~\citep{chawla2002smote}, to teaching large language models (LLMs) to write code from synthetically generated textbooks~\citep{gunasekar2023textbooks}, synthetic data has enabled researchers and engineers to create useful approximations of real-world data that would otherwise be scarce, costly, sensitive, dangerous, or impossible to obtain, while providing greater control over the scale, composition, and characteristics of the resulting datasets.

% pre- and post-training
Today, SDG plays a central role across the entire lifecycle of LLMs and the agentic systems they power. At the pretraining stage, where high-quality human-written text is not keeping pace with demand~\citep{villalobos2024data}, synthetically rephrased and augmented web corpora commonly contribute tens of billions to trillions of tokens~\citep{maini2024wrap,benallal2024cosmopedia,su2025nemotroncc}. Large-scale studies of this regime show that rephrased text reduces the training tokens needed to reach a given loss when mixed with natural text but not when used alone~\citep{kang2025demystifying}, and that the choice and diversity of the rephrasing prompts strongly affect downstream performance~\citep{finephrase2026}. Post-training, in turn, leverages synthetic instructions, preferences, and reasoning traces to teach models new behavior~\citep{wang2023selfinstruct, taori2023alpaca,xu2023wizardlm,honovich2023unnatural}. Studies of filtered instruction sets, constrained-vocabulary stories, and textbook-quality code show that the composition and quality of such data can matter more than its volume~\citep{zhou2023lima,chen2023alpagasus,liu2024deita,eldan2023tinystories,gunasekar2023textbooks}. 

% agents and multimodal
Teaching models agentic and multimodal capabilities also heavily relies on SDG. For multi-step tasks, such as searching the web~\citep{nakano2021webgpt}, synthetic interaction trajectories are generated at scale~\citep{qin2024toollm,kimi2025k2}. For visual understanding, textual annotations are generated for images, documents, and videos, and increasingly non-text modalities themselves are synthesized as well~\citep{yang2025cosyn,chen2025longvila,nvidia2026nemotron3nanoomni}. The Nemotron~3 model reports describe synthetic data pipelines for code, STEM reasoning, structured-output adherence, SQL, search, tool use, document extraction, and citation behavior~\citep{nvidia2025nemotron3nano,nvidia2026nemotron3super,nvidia2025nemotron3ultra}.

% the gap
The emerging theme is that the usefulness of synthetic data depends on deliberate control over its composition and diversity~\citep{kang2025demystifying,finephrase2026}, verification of its correctness and quality~\citep{feng2024verification,chen2023alpagasus}, and its mixture with and grounding in real-world data~\citep{gerstgrasser2024modelcollapse}. In practice, SDG pipelines typically accomplish this using bespoke scripts and glue code, which makes them difficult to interpret, share, and reproduce.

\begin{figure}[t]
\centering
\includegraphics[width=0.98\textwidth]{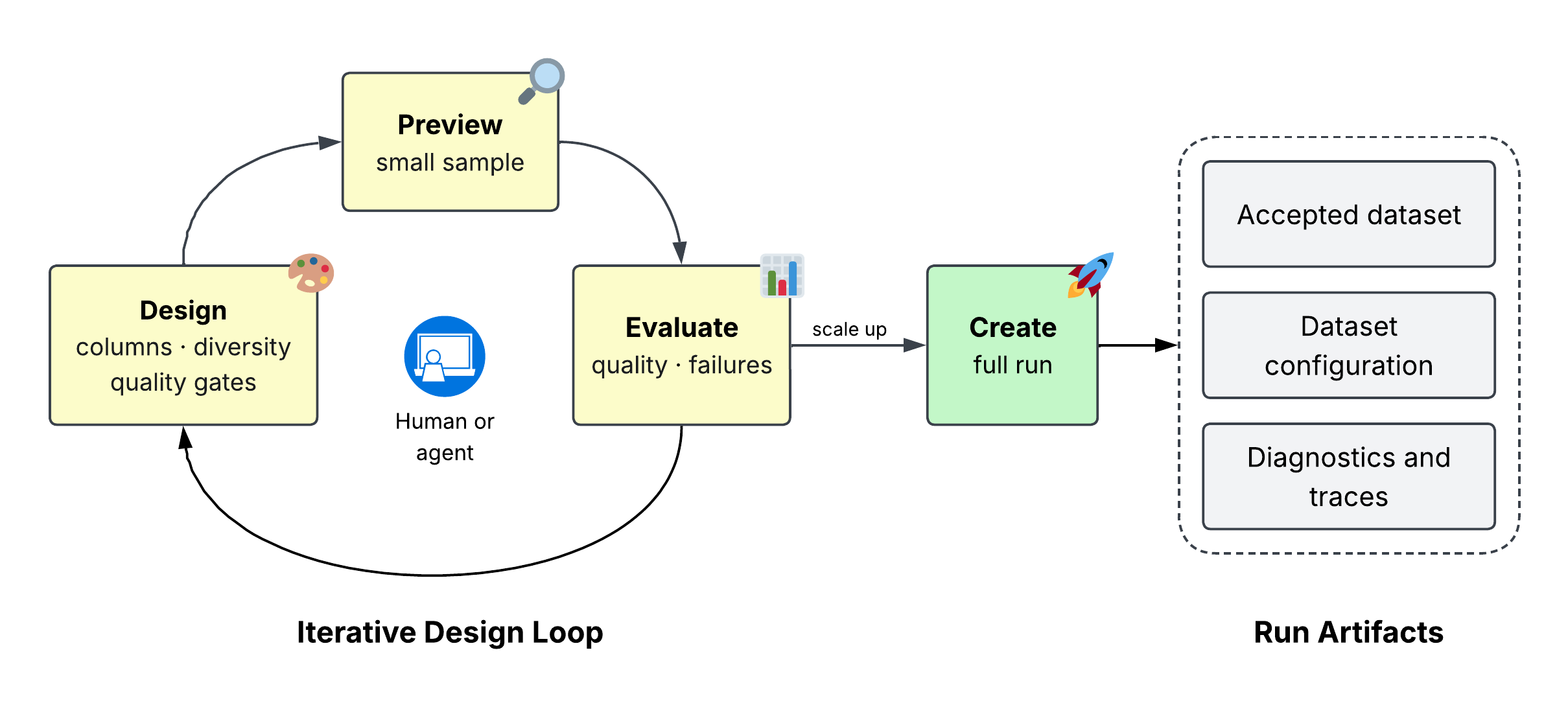}
\caption{\textbf{Typical NeMo Data Designer user workflow.}
Define dataset columns, generate and evaluate a small preview sample, and iterate until satisfied. A full run then creates the dataset from the same configuration and saves it along with useful run artifacts.}
\label{fig:dd-hero}
\end{figure}

% intro NDD
In this report, we present \library(\ndd),\footnote{Code available at \url{https://github.com/NVIDIA-NeMo/DataDesigner}.} an open-source, general-purpose framework for multimodal SDG. \ndd is designed to be intuitive for both humans and agents. Datasets are specified in a declarative configuration, which is itself an inspectable, shareable artifact that makes the generation pipeline reproducible. Built-in column types include model-generated text, code, structured outputs, images, and embeddings. Sampler columns draw from statistical distributions to steer diversity, and validator and judge columns verify the generated values. New column types can be added through a flexible plugin system, and by combining built-in and custom columns, most of the generation strategies cited above can be expressed within \ndd.

SDG is an inherently iterative process, since dataset requirements are rarely met on the first generation attempt. \ndd therefore builds iteration into its core design. We illustrate the typical user workflow in Figure~\ref{fig:dd-hero}. In the iterative design loop, the user declares the columns of the dataset, generates and reviews a small preview sample, updates the configuration based on their observations, and repeats until satisfied. The full dataset is then generated at scale from the same configuration. At runtime, \ndd resolves dependencies between columns, schedules calls to user-provided model endpoints, and retries failed requests. Synthetic datasets built with \ndd have been used in Nemotron model development~\citep{nvidia2025nemotron3nano,nvidia2026nemotron3super,nvidia2026nemotron3nanoomni,nvidia2025nemotron3ultra} and in production enterprise deployments spanning structured, agentic, multimodal, and domain-specialized tasks (Section~\ref{sec:applications}).

The key contributions of this work are the following: 

\begin{itemize}[itemsep=1.5ex]
    \item \textbf{An open-source, general-purpose framework for multimodal SDG.} \ndd replaces one-off pipeline scripts with a declarative configuration that is intuitive for both humans and agents. The same specification drives preview, iteration, and full-scale generation, making SDG pipelines easy to interpret, share, and reproduce.
    \item \textbf{Explicit statistical control over dataset diversity.} Statistical samplers are first-class column types, with values that follow user-specified distributions and serve as inputs to generation prompts, giving users direct control over dataset composition and diversity.
    \item \textbf{Extensibility.} A flexible plugin system allows new column types and functionality to be added without modifying the core framework, making it possible to implement novel generation strategies and explore new modalities in \ndd.
    \item \textbf{Real-world impact from research and production.} We highlight datasets built with \ndd for structured, agentic, multimodal, and domain-specialized tasks, including data used in Nemotron model development and in enterprise deployments, and summarize their reported downstream results.
\end{itemize}

The remainder of this report is organized as follows. Section~\ref{sec:architecture} describes \ndd's architecture. Section~\ref{sec:methodology} presents the pipeline design methodology that recurs across applications. Section~\ref{sec:applications} describes representative workflows and case studies together with their evaluation results. Section~\ref{sec:conclusion} concludes, followed by limitations, broader impact, and availability.

% This section uses lstlisting from the listings package.
\section{Architecture}
\label{sec:architecture}

\begin{figure}[!t]
\centering
\includegraphics[width=0.88\textwidth]{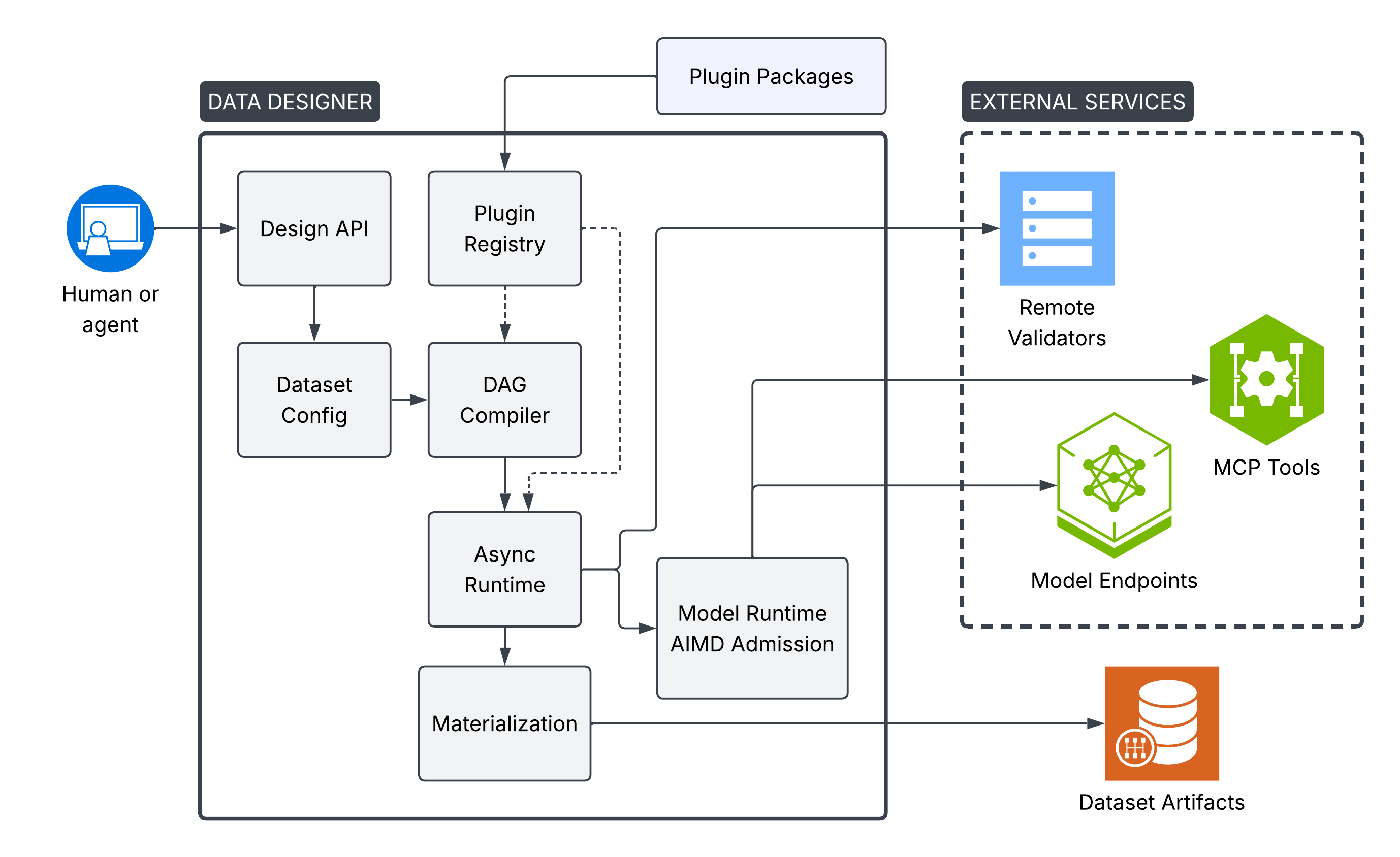}
\caption{\textbf{Component architecture and external-service boundaries.}
Solid arrows show the primary configuration and execution path; dashed arrows
show plugin extension points. Model endpoints, MCP tools, and remote validators
remain outside \ndd's process boundary.}
\label{fig:dd-architecture}
\end{figure}

Behind that user-facing workflow, \ndd keeps dataset design, execution, and external services separate. Figure~\ref{fig:dd-architecture} shows how those pieces fit together and where the system boundary sits.

\subsection{Overview}

The design API collects declarative definitions for columns and their supporting models, tools, seeds, processors, and profilers. Installed plugin packages are discovered through the plugin registry; their configuration types participate in compilation, and their implementations enter the same execution paths as built-in components. The DAG compiler validates the configuration and resolves it into a directed acyclic graph of columns. Each column produces one named value, which may appear in the final dataset or serve as input to downstream columns. Some columns also emit auxiliary side-effect fields, such as message traces. Dependencies are inferred from explicit configuration and from Jinja2 references in prompts and expressions, so a column that uses \texttt{\{\{ schema \}\}} or \texttt{\{\{ persona.age \}\}} automatically waits for those values to exist. Circular dependencies are rejected when the configuration is compiled, before generation begins.

The asynchronous runtime schedules ready work from this graph and routes
model-backed tasks through the model runtime, where adaptive request admission
regulates calls to external endpoints. MCP tools and remote validators are
invoked across explicit service boundaries. Materialization writes completed
row groups, traces, processor views, and final dataset artifacts without making
the external services part of the declarative graph.

This separation keeps dataset design, model selection, validation policy, and
execution machinery independently visible. The same configuration can therefore
describe a dataset without assuming where model weights are hosted or how
endpoint capacity is provisioned.

\subsection{Design Interface}

Dataset designs can be written as declarative YAML or constructed in Python with the \texttt{DataDesignerConfigBuilder}. \texttt{preview()} and \texttt{create()} use default execution settings, while an optional \texttt{RunConfig} lets users override those settings when needed. Keeping execution settings outside the column graph allows the same dataset design to be previewed and run at full scale. Appendix~\ref{app:declarative-config-example} shows the same design in YAML and Python.

Across both interfaces, fields are declared independently and their dependencies are encoded in data rather than in imperative orchestration code. References such as \verb|{{ audience }}| and \verb|{{ goal }}| identify both prompt substitutions and upstream dependencies, allowing \ndd to compile the specification into an executable column graph. The workflow author declares what each field requires, while the runtime determines when the corresponding work is ready to execute.

Model-backed columns follow the same pattern: they name a logical model alias without embedding endpoint, credential, or serving details in the column definition. Section~\ref{sec:models-and-providers} describes how these aliases resolve into the runtime model stack.

\begin{figure}[t]
\centering
\includegraphics[width=\textwidth]{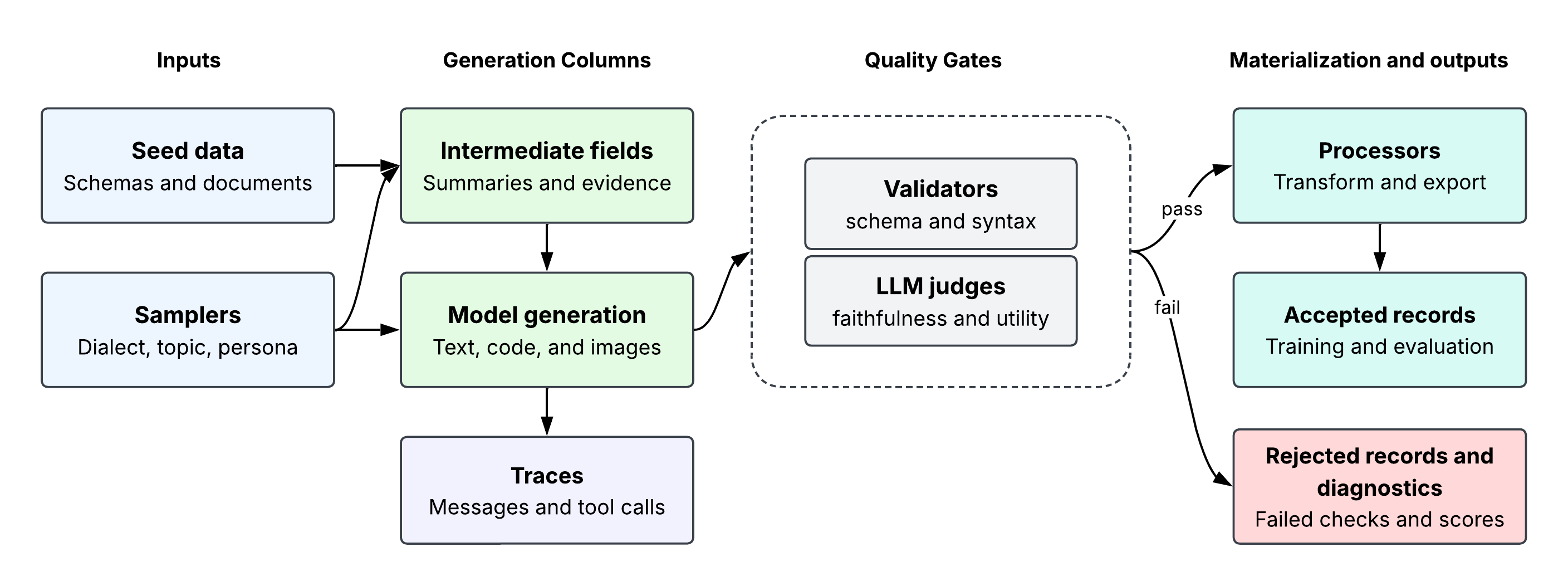}
\caption{\textbf{Dependency-aware generation, quality gating, and materialization.}
Seed data and samplers establish record inputs and controlled variation.
Dependent columns construct intermediate fields and model-generated candidates,
while validators and LLM judges separate passing records for transformation and
export from rejected records with inspectable failure diagnostics. Model
interactions can also be retained as message and tool-call traces.}
\label{fig:dd-column-dag}
\end{figure}

\subsection{Columns as Pipeline Building Blocks}

Figure~\ref{fig:dd-column-dag} summarizes the dependency-aware path from record inputs to materialized outputs. Seed and sampler columns establish the record population and controlled variation; intermediate and model-backed columns construct candidate records in stages; and validators and LLM judges apply structural and semantic quality gates. Passing records can be reshaped by processors for training or evaluation, while rejected records retain failed checks and scores for diagnosis. Message and tool-call traces remain separate, inspectable side effects of model generation. Keeping these stages distinct exposes the pipeline as a sequence of design choices rather than a single overloaded prompt.

This vocabulary lets a workflow author decompose generation into inspectable stages. A retrieval pipeline can separate document chunking, artifact extraction, question generation, embedding-based deduplication, judge scoring, and export. A text-to-SQL pipeline can separate schema sampling, prompt synthesis, SQL generation, dialect validation, and rubric scoring. Failures are consequently attached to a stage instead of being hidden inside one prompt. Coverage planning is discussed in Section~\ref{sec:sampling-and-staging}.

\subsection{Models and Providers}
\label{sec:models-and-providers}

Model aliases extend \ndd's declarative programming model to inference. A model-backed column names the logical model role it requires rather than an endpoint. A pipeline can therefore use separate aliases for creative generation, lower-variance judging, and embeddings, even when some aliases target the same underlying model. An alias can also be rebound from a hosted API to an enterprise gateway or self-hosted endpoint without changing prompts or the dependency graph.

\ndd maintains this separation through two configuration objects. A \texttt{ModelProvider} describes the serving boundary, including how and where requests are sent. A \texttt{ModelConfig} binds a workflow-visible alias to that provider, the underlying model identifier, the operation family (chat, embedding, or image), and an inference policy. The provider therefore answers \emph{where and how a request is sent}, while the model configuration answers \emph{which model role and policy a column uses}.
Model choice and inference policy remain part of the inspectable dataset specification, while credentials and endpoint topology remain provider concerns.

Listing~\ref{lst:appendix-model-provider-alias-config} in Appendix~\ref{app:config-examples} gives a complete provider-and-alias example, including concrete parameter classes and two aliases that apply different inference policies to the same underlying model.

Before a run begins, \ndd resolves only the aliases referenced by the compiled graph and verifies that their endpoints are available, unless health checking has been disabled. This catches unavailable model dependencies before generation starts and avoids constructing clients that the workflow never uses.

At runtime, each alias resolves to a shared model facade for chat completion, embeddings, and image generation. Provider adapters handle service-specific protocols and errors, while an admission controller adapts concurrency to endpoint rate limits. Appendix~\ref{app:model-runtime} describes this runtime path in detail.

\subsection{Async Runtime}
\label{sec:async-runtime}

The compiled DAG establishes what depends on what; the asynchronous runtime decides when each unit of work can proceed. In a heterogeneous pipeline, stages rarely finish at the same pace, so a column-at-a-time schedule would leave ready work waiting for the slowest rows. \ndd instead treats the DAG as a live execution plan. As soon as the required inputs for a row exist, the dependent cell becomes eligible to run. If \texttt{summary} and \texttt{trivia} both depend on \texttt{topic}, they can run together, and an \texttt{analysis} field can begin for each row as soon as its \texttt{summary} is ready. Independent branches therefore overlap without changing the declarative configuration.

The runtime processes records in row groups, which bound memory use and provide the unit of checkpointing. Within a row group, readiness is tracked at the granularity appropriate to each column. Model-backed text, embedding, and image columns usually create one task per row, while samplers, seed readers, and vectorized custom columns operate on the group as a whole. A row-group task can wait for its upstream inputs without preventing unrelated cell-level work from advancing. Columns may also be skipped conditionally for individual rows. When that happens, only downstream work that depends on the skipped value is affected; unrelated branches continue.

When many branches become ready together, the scheduler uses a fair queue and enforces global, group, and resource limits. Work is grouped by provider, model, and request domain, or by local generator, so one wide or rate-limited branch does not dominate unrelated work. Model calls pass through the adaptive request-admission layer detailed in Appendix~\ref{app:model-runtime}. Each engine invocation remains a single asynchronous process; partitioning a run across processes belongs to the deployment layer described in Section~\ref{sec:execution-and-deployment-boundary}.

For long-running jobs, row groups also make progress durable. Completed groups can be checkpointed in Parquet and released from memory while later work continues, even when groups finish out of order. If a run is interrupted or fails, a later invocation can resume from those checkpoints; before continuing, the engine reconstructs progress and checks configuration compatibility. Retryable failures follow the configured retry policy, while permanent cell failures remove the affected row and its downstream work. If the non-retryable error threshold stops a run early, completed rows are preserved and partial rows are discarded. Optional task traces record dispatch, admission, completion, status, and errors, helping distinguish queueing delay from execution time during diagnosis.

\subsection{From Generation to Usable Data}

Execution does not end when the runtime produces a value. \ndd must also determine whether that value is usable and preserve enough context to explain how it was produced. Validation therefore appears directly in the column graph: deterministic validators, remote services, and LLM judges consume generated fields and add pass/fail metadata or scores. Because these quality gates are graph stages, the scheduler can place them after the data they inspect while keeping validation policy separate from the original generation prompt. Remote validators cross the external-service boundary shown in Figure~\ref{fig:dd-architecture}.

Alongside these quality gates, model-backed columns can call MCP tools when they need outside information or actions. Each tool result returns to the ongoing model interaction, and the message history can be retained as a trace rather than mixed into the generated field. After generation and validation, processors reshape batches or completed datasets, and materialization gathers the primary rows, processor-derived views, traces, checkpoints, and diagnostics into inspectable artifacts. Together, these stages turn raw generations into data that can be evaluated, audited, and reused.

\begin{figure}[!t]
\centering
\includegraphics[width=0.96\textwidth]{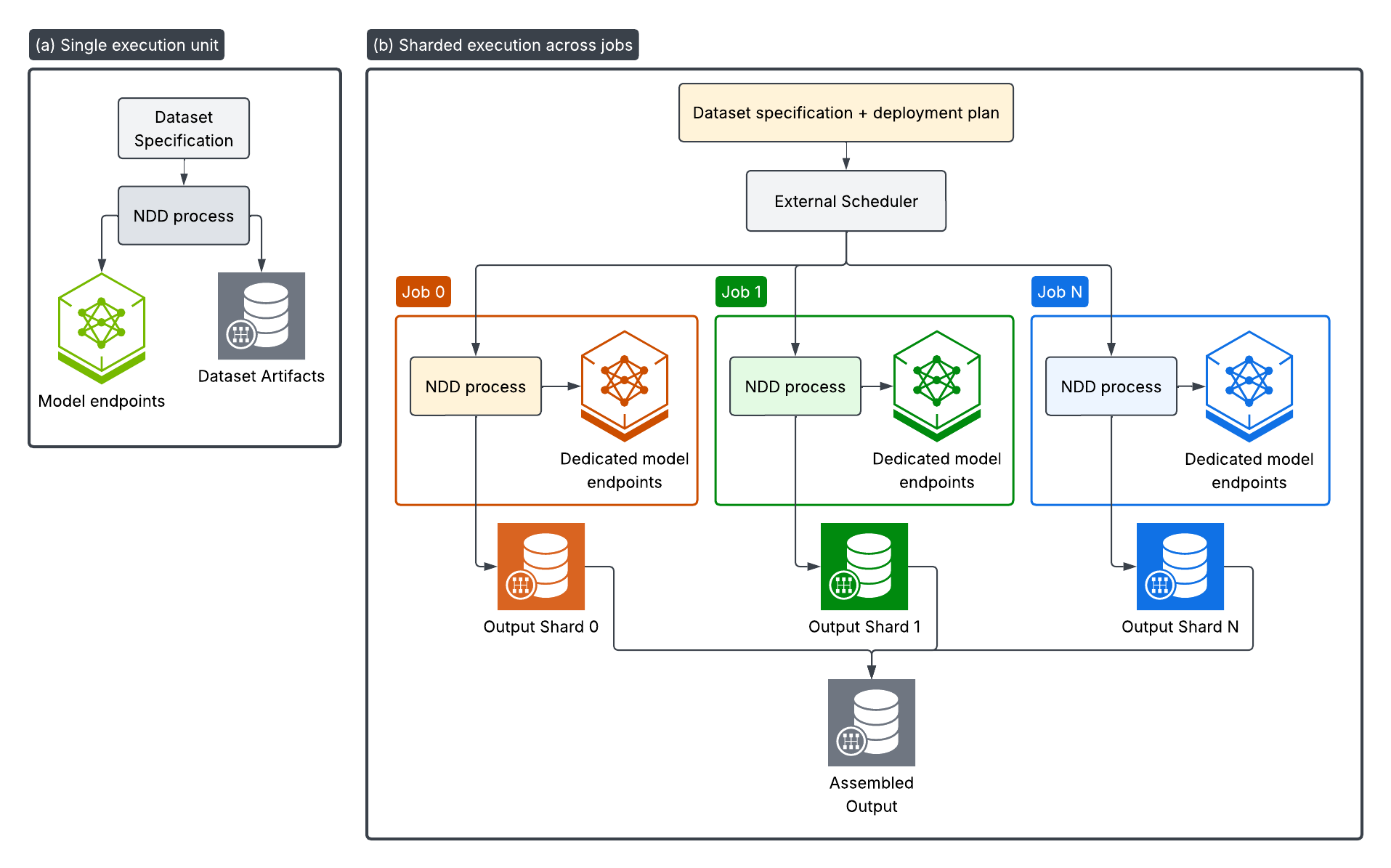}
\caption{\textbf{Single-job and sharded deployment patterns.}
Panel (a) runs one \ndd process against model endpoints and materializes one
artifact set. In panel (b), a dataset specification and deployment plan are
submitted to an external scheduler, which launches independent jobs with
dedicated endpoint deployments and color-matched output shards. A downstream
assembly step merges those shards into the final dataset.}
\label{fig:deployment-scale-out-patterns}
\end{figure}

\subsection{Trust and Deployment Boundaries}
\label{sec:trust-boundaries}
\label{sec:execution-and-deployment-boundary}

Two boundaries matter once a pipeline reaches beyond its local data flow: the
trust boundary governs what may execute or leave the process, while the
execution boundary defines what \ndd coordinates directly.

Model endpoints, remote tools, executable validators, and installed plugins are explicit trust boundaries. Prompts, seed snippets, intermediate fields, multimodal inputs, tool schemas, and tool observations may leave the local process when sent to a hosted endpoint, gateway, remote tool, or validation service. Sensitive workflows should use approved endpoints, external secret management, and redaction before egress, and should record provider logging, retention, and geographic-processing assumptions. Because these services can change independently of the dataset specification, they can also prevent byte-for-byte regeneration. Section~\ref{sec:export-reproducibility} therefore treats reproducibility as an auditable recipe that makes service-dependent variation visible.

Executable code and SQL, local callables, remote validators, and plugin packages should run with isolation, scoped credentials, timeouts, network and filesystem limits, and audit logging appropriate to the data. Tool-enabled workflows should additionally allowlist MCP providers and tools and review or redact captured traces before training or publication. Consolidating these controls at the service boundaries keeps them separate from the data-flow semantics of validators, traces, and processors.

Within its execution boundary, \ndd compiles the dataset configuration,
schedules ready work, batches requests, applies retries and validation, adapts
request admission, and materializes checkpoints and outputs. Provisioning,
multi-process partitioning, and distributed orchestration remain external
responsibilities.

For larger runs, a deployment plan assigns deterministic shards and an external
scheduler launches one independent \ndd process per shard. Each process
maintains its own scheduler, adaptive request-admission state, checkpoints,
traces, and artifact namespace. The serving topology remains a deployment
choice: a job may provision dedicated model endpoints, as shown in
Figure~\ref{fig:deployment-scale-out-patterns}, or multiple jobs may share
externally managed endpoints. Each job writes a separate shard artifact, which
an external assembly step merges into the final dataset.
Appendix~\ref{app:operational-deployment-and-capacity-planning} provides
capacity-planning guidance.

\FloatBarrier
\subsection{Plugins}

Scaling out changes how many independent runtimes execute a configuration. Plugins extend what that configuration can declare. Dataset projects often need an internal corpus layout, simulator, domain library, redaction policy, or trainer schema that does not fit a built-in component. \ndd brings this specialized behavior into the declarative workflow so it remains previewable and auditable.

The plugin system is the supported extension boundary for this behavior. It identifies three plugin types: seed readers bring new source systems into the record population, column generators produce values with the same dependency semantics as built-in columns, and processors transform batches or final datasets into project-specific inspection, training, or evaluation views. One-off custom columns remain useful for prototyping, but a plugin is the better boundary when logic needs a stable configuration schema or reuse across projects.

A plugin package keeps the same separation of concerns as the core framework. A user-facing configuration class declares options, discriminators, dependencies, and metadata; an implementation class performs the narrow runtime behavior; a plugin descriptor connects the two; and a Python entry point exposes that descriptor after installation. Users can then import the plugin's configuration classes and continue using \texttt{DataDesignerConfigBuilder}, \texttt{preview()}, and \texttt{create()} as before. Column-generator plugins become graph nodes with normal dependency semantics, while seed readers and processors use the same declarative workflow and runtime paths.

Plugins can be distributed as independent Python packages, complete with their own dependencies, documentation, tests, and ownership metadata, through first-party, internal, or community catalogs. \ndd still owns dependency resolution, scheduling, model-client integration, validation, traces, and output materialization. Appendix~\ref{app:plugin-example} gives a minimal implementation and usage example.

\section{Pipeline Design}
\label{sec:methodology}

\ndd pipelines differ by domain, but successful workflows tend to follow the same compact recipe: define the record contract, make the desired variation explicit, preview examples before scaling, validate aggressively, and preserve enough provenance to understand both accepted and rejected data. We use the term \emph{statistical grounding} to indicate that important sources of variation--topic, task type, difficulty, persona, schema shape, tool budget, output format, or document provenance--are represented as seed or sampler fields before the LLM is asked to generate content. The dataset is therefore shaped by a visible design plan rather than by a single prompt asking for diversity.

\subsection{Design Loop}

The design loop begins with the record contract: the fields to generate, the downstream trainer or evaluator format, and the labels or traces that must survive export. The designer then chooses seed sources and coverage axes such as documents, schemas, personas, topics, difficulty levels, document layouts, or tool environments. These choices are encoded as seed-dataset and sampler columns so they can be inspected before any large model run. Generation is decomposed into stages, with model calls reserved for semantic work and deterministic transformations used for parsing, reshaping, or bookkeeping. Quality gates are specified early, including parsers, validators, dialect or schema checks, LLM judges, distribution checks, and any human review needed for high-risk outputs.

\texttt{preview()} is the main checkpoint in this loop. A small run should expose sampled variables, intermediate fields, rejected examples, and the final record shape. The preview step answers practical questions before an expensive \texttt{create()} run: whether the sampled attributes match the intended distribution, whether prompts receive the right context, whether generated records are repetitive, whether validators reject the intended failures, and whether the output schema matches the downstream consumer. After a full run, evaluation results and rejected-record diagnostics feed the next iteration.

\subsection{Sampling and Staging}
\label{sec:sampling-and-staging}

Seed curation determines what generation will amplify. Seeds may be documents, database schemas, code snippets, agent rollouts, questions, personas, product catalogs, media assets such as images, audio recordings, and videos, or unlabeled domain records. Before adding generation columns, the designer should decide which seed attributes become labels, which become conditioning context, which are only diagnostics, and which must remain visible after export. For document-grounded retrieval, segment identifiers must stay attached to generated questions so positive passages survive filtering and export. For science reasoning, stratified topic and FDC-code sampling prevents overrepresented domains from dominating the generated corpus. For enterprise query workflows, deduplication, privacy scrubbing, and schema normalization often belong before generation begins~\citep{crowdstrike2026}.

Sampler columns turn a coverage plan into explicit inputs. Category and
subcategory samplers support weighted and hierarchical choices; numerical and
SciPy samplers cover common probability distributions; conditional sampling
changes parameters based on earlier columns; and person samplers can create
Faker-style profiles or draw from richer Nemotron-Personas assets. These
operations are cheaper and more predictable than model calls, so they shape
dataset composition before generation begins. Constraints extend this control
from marginal to joint conditions: a sampler column can be bounded by an
inequality against a scalar or another sampler column, allowing orderings and
thresholds to hold by construction rather than through generate-and-discard
filtering that can skew the accepted distribution.

Controlled sampling turns a desired coverage plan into executable conditions. A simple dataset may sample language, style, or difficulty; a more complex one may sample database dialect, schema size, dirty-data pattern, distractor count, user persona, output format, hop count, evidence window, or tool budget. Because these variables are ordinary columns, the accepted dataset can be checked against the intended mix after validation. This also reduces pressure on prompts: instead of asking a model to ``be diverse,'' the pipeline supplies concrete conditions such as a schema with distractor columns, a strict JSON schema, or a physics passage from an underrepresented topic.

Most useful records are assembled through several stages. Retrieval workflows can extract document concepts, generate questions, write answers, deduplicate, and judge grounding. Text-to-SQL workflows can sample a domain and dialect, synthesize schema context, generate a query, then apply syntax and semantic checks. Tool-use workflows can construct a seed question, roll out a model through an MCP tool interface, capture the trajectory, and convert successful traces into supervised examples. Document-understanding workflows can select
pages or page windows, attach rendered images and extracted text as context, generate
questions targeted at a reasoning mode, and filter with a vision-language judge. Staging keeps each failure local enough to debug and allows stronger or more expensive models to be reserved for the steps that need them.

\subsection{Validation and Provenance}

Validation converts candidates into usable data. The right gate depends on the target artifact: parse and schema checks for JSON, YAML, XML, or function-call data; code and SQL validators for executable or dialect-constrained outputs; grounding checks for document or tool-supported answers, including vision-language judging when the evidence is an image or page; LLM judges for semantic criteria such as relevance, faithfulness, or difficulty; and post-filter distribution checks when rejection may skew coverage. Rejection sampling is the common operational pattern: generate more candidates than needed and keep the records that pass the gates; when filtering skews the intended mix, coverage is rebalanced by curating seeds and samplers rather than by a single prompt. The workflows in this report use this pattern for structured-output examples, text-to-SQL records, and search-agent trajectories.

Provenance matters because synthetic data is selected, not merely generated. During development, intermediate columns, seed identifiers, prompt inputs, model aliases, validation scores, and dropped examples are often as useful as the final record. For agentic data, full traces are part of the training signal: the sequence of searches, tool results, intermediate turns, and final answer teaches behavior that a standalone answer cannot. Preserving these traces also makes failures inspectable when a trajectory searches for the wrong entity, ignores evidence, exceeds a tool budget, or reaches the right source but synthesizes the wrong response.

\subsection{Export and Reproducibility}
\label{sec:export-reproducibility}

The final stage reshapes accepted records for downstream use. Export and processor stages can render accepted records as chat-format messages, prompt/completion views, additional parquet shards, evaluator schemas, or inspection views without changing the source generation columns; the Retriever SDG plugin further exports BEIR-style retrieval artifacts. The same generated data may therefore support training, evaluation, manual audit, and public release. Data Designer can publish these artifacts directly to the Hugging Face Hub through \texttt{results.push\_to\_hub()}. Alongside the dataset and processor-derived views, the upload includes \texttt{builder\_config.json}, a serialized representation of the dataset's declarative design state that can be inspected, reloaded, revised, and rerun.

Reproducibility is not perfect determinism; model endpoints, provider settings, and external tools may change. The practical target is an auditable recipe: seed identifiers, sampler parameters, prompt templates, model aliases and versions where available, inference settings, validation thresholds, judge rubrics, tool configurations, trace settings, processor outputs, and export schemas. Keeping rejected examples and intermediate fields is often the fastest way to understand why a dataset changed between iterations.

% Combined replacement for use_cases.tex and case_studies.tex.
% Expected preamble packages already used by the report include:
% graphicx, tabularx, booktabs, ragged2e, listings, needspace, url, and natbib.
% Upload the three image files to figures/ in Overleaf.

\section{Applications and Case Studies}
\label{sec:applications}
% Compatibility label for references that previously pointed to Section 6.
\label{sec:evaluation}

An SDG framework is ultimately tested by the datasets it can produce and validate in practice. Within the Nemotron program, \ndd has been used to create targeted data for specific model capabilities. \ndd has also supported enterprise applications and external collaborations, translating partner-specific data gaps into validated datasets under domain, privacy, validity, and production constraints. Together, these applications demonstrate the framework in settings where generated records must survive task-specific filtering and remain usable for downstream training and evaluation.

The section progresses from tightly verifiable output and prompt constraints, through executable query generation and multi-step search trajectories, to multimodal document understanding and statistically grounded personas. Each case addresses the same questions: What capability or data gap motivated the workflow? What data was required? How did \ndd generate and validate it? Where were the accepted records used, and what outcome was observed? Table~\ref{tab:applications-summary} maps each application to the primary \ndd strength it demonstrates and the evidence presented.

\begin{table*}[t]
\centering
\footnotesize
\setlength{\tabcolsep}{4pt}
\renewcommand{\arraystretch}{1.08}
\begin{tabularx}{\textwidth}{@{}
>{\RaggedRight\arraybackslash}p{0.17\textwidth}
>{\RaggedRight\arraybackslash}p{0.17\textwidth}
>{\RaggedRight\arraybackslash}p{0.22\textwidth}
>{\RaggedRight\arraybackslash}X@{}}
\toprule
\textbf{Application} &
\textbf{Primary \ndd strength} &
\textbf{Data scale} &
\textbf{Reported outcome} \\
\midrule

Structured outputs and model usability &
Programmatic verification &
9,949 verified JSON examples and additional Nemotron 3 Ultra usability data &
JSONSchemaBench: 80.2\% $\rightarrow$ 86.9\%; StructEval-Text: 64.5\% $\rightarrow$ 72.1\%; Ultra StructEval-T: 78.6\% $\rightarrow$ 82.1\% \\
\addlinespace[1mm]

Text-to-SQL and CrowdStrike CQL &
Staged structural generation and reverse labeling &
300K SQL candidates with 96.5K retained; privacy-scrubbed CrowdStrike CQL queries paired with generated descriptions &
BIRD: 26.77\% $\rightarrow$ 41.80\% (GPT-OSS-120B: 38.25\%); CrowdStrike model: 96\% valid-query accuracy and 2.50/5 semantic score \\
\addlinespace[1mm]

Search-agent trajectories &
Tool-trace preservation &
50K seeds $\rightarrow$ 24K questions $\rightarrow$ approximately 7K valid trajectories; approximately 12 tool calls per accepted trajectory &
Validated multi-turn trajectories were incorporated into Nemotron 3 Super supervised fine-tuning \\
\addlinespace[1mm]

Long-document VLM understanding &
Iterative multimodal capability targeting &
Approximately 11.4M visual question-answer pairs, representing about 45B tokens &
MMLongBench-Doc development checkpoints: 26.32\% $\rightarrow$ 59.00\%; released Nemotron 3 Nano Omni: 57.5\% \\
\addlinespace[1mm]

Prompt robustness &
Controlled linguistic diversity &
50 seed combinations expanded across six variation axes &
Approximately 2$\times$ lower prompt-sensitivity variation across GPQA, MMLU-Pro, competition mathematics, and LiveCodeBench \\
\addlinespace[1mm]

Nemotron-Personas &
Population-grounded diversification &
10 regional datasets across 15 language or script editions; regions representing approximately 2.4B people &
Reused in Nemotron training and independently adopted for post-training, safety testing, and evaluation \\
\bottomrule
\end{tabularx}
\caption{Representative \ndd applications, the framework strength each exposes most directly, the scale of the generated data, and the reported outcomes.}
\label{tab:applications-summary}
\end{table*}

\subsection{Structured Outputs and Model Usability}
\label{sec:app-structured-outputs}

\paragraph{Programmatic verification.}
Reliable structured outputs allow models and agents to call tools, populate APIs, and pass machine-readable state between workflow steps. Training data for this capability pairs diverse instructions and grounding content with record-specific schemas and responses that conform to them. \ndd samples topics, schema complexity, output format, and prompt arrangement; generates a schema and grounding document; produces multiple candidate responses; and applies format-specific parsing and schema validation before selecting an accepted rollout. Schema conformance is therefore checked programmatically rather than inferred from an LLM judge. Details of the pipeline, released dataset, and results are provided in the structured-output developer note.\footnote{\url{https://docs.nvidia.com/nemo/datadesigner/dev-notes/structured-outputs-from-nemotron}}

Nemotron 3 Nano used \ndd to create approximately 9K JSON-schema adherence tasks for reinforcement learning, with reward determined by exact schema conformance~\citep{nvidia2025nemotron3nano}. The corresponding released dataset contains 9,949 verified JSON examples. Training Nemotron Nano v3 with this data improved accuracy from 80.2\% to 86.9\% on JSONSchemaBench and from 64.5\% to 72.1\% on StructEval-Text~\citep{geng2025jsonschemabench,yang2025structeval}. Although the released dataset is JSON-only, the StructEval-Text results show improvements across output formats, with TOML and XML remaining the weakest.

\paragraph{Enterprise Case Study: Nemotron 3 Ultra Usability for Perplexity AI}

Perplexity's search engine depends on strict output formats and correct inline citations. Data Designer turned those product requirements into verifiable training data for Nemotron 3 Ultra. Building on the structured-schema capabilities developed for Nemotron 3 Super, the Nemotron 3 Ultra technical report describes a usability curriculum spanning JSON, YAML, XML, TOML, and CSV; document extraction with complex nested fields and distractors; multiple inline citation formats; and free-form responses that follow detailed formatting instructions.\footnote{\url{https://research.nvidia.com/labs/nemotron/files/NVIDIA-Nemotron-3-Ultra-Technical-Report.pdf}} \ndd generated seed data covering schema adherence, answer formatting, and single-source, multi-source, and Perplexity-style inline citations. Deterministic verifiers checked schema validity, formatting compliance, reference correctness, and citation coverage. The resulting environments were incorporated into Ultra's RLVR and model-usability teacher training.

Following training that included these datasets, Nemotron 3 Ultra's StructEval-T accuracy improved from 78.6\% to 82.1\%. This work formed part of the broader NVIDIA--Perplexity collaboration through the Nemotron Coalition, to which Perplexity contributed data and evaluations.\footnote{\url{https://www.perplexity.ai/hub/blog/perplexity-joins-the-nvidia-nemotron-coalition}}

\subsection{Text-to-SQL}
\label{sec:app-text-to-sql}

\paragraph{Staged structural generation.}
Text-to-SQL data trains a model to translate natural-language requests into executable, dialect-appropriate queries grounded in realistic database schemas. Useful records must jointly preserve the request, schema and sample data, SQL dialect, target query, reasoning trace, and validation metadata. \ndd first samples industry, domain, SQL concept, difficulty, dirty-data pattern, task type, and linguistic style; it then generates the request, schema context, and SQL query in dependent stages. Schema generation deliberately introduces distractor tables and columns and production-like irregularities such as dates stored as text, currency symbols, and embedded JSON. Dialect-specific validators and five LLM critics score syntax and semantic qualities before filtering. Details are provided in the developer note and enterprise recipe.\footnote{\url{https://docs.nvidia.com/nemo/datadesigner/dev-notes/text-to-sql-for-nemotron-super}}\footnote{\url{https://github.com/NVIDIA-NeMo/DataDesigner/blob/main/docs/assets/recipes/code_generation/enterprise_text_to_sql.py}}

\begin{figure}[!t]
\centering
\includegraphics[width=0.80\linewidth]{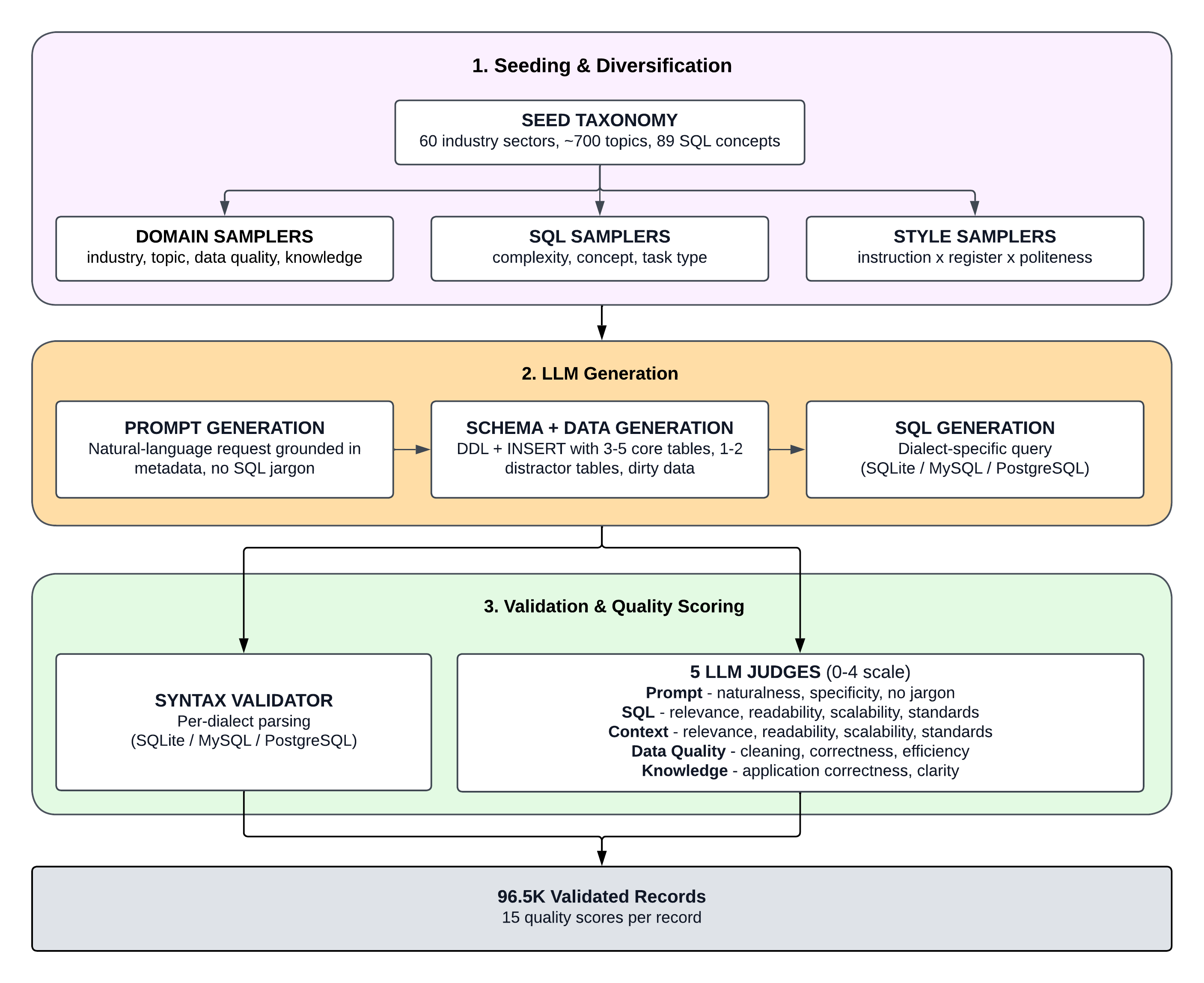}
\caption{\textbf{Text-to-SQL synthetic-data pipeline.}
A seed taxonomy and controlled samplers establish domain, SQL, and stylistic
variation. Dependent LLM stages generate the request, schema and data, and
dialect-specific query before per-dialect validation and five LLM judges
produce quality signals for record selection.}
\label{fig:text-to-sql-workflow}
\end{figure}

Figure~\ref{fig:text-to-sql-workflow} summarizes the pipeline from controlled
seeding through dependent generation, validation, and record selection.

The workflow generated 300K candidates across PostgreSQL, MySQL, and SQLite and retained 96.5K after rejecting approximately 68\% through its quality waterfall. Listing~\ref{lst:text-sql-record} illustrates the retained record shape.

\begin{lstlisting}[float=!tb,caption={Illustrative text-to-SQL record after filtering and reshaping.},label={lst:text-sql-record}]
{
  "messages": [
    {"role": "user", "content": "List paid invoices."},
    {"role": "assistant", "content": "SELECT invoice_id FROM invoices WHERE status = 'paid';"}
  ],
  "sql_context": "tables: invoices(invoice_id, customer_id, status)",
  "metadata": {
    "sql_dialect": "PostgreSQL",
    "sql_complexity": "Intermediate",
    "is_valid": true,
    "sql_relevance_score": 4,
    "provenance_columns": [
      "sql_dialect", "sql_context", "sql",
      "is_valid", "sql_relevance_score"
    ]
  }
}
\end{lstlisting}

\paragraph{Enterprise case study: CrowdStrike CQL.}
CrowdStrike used \ndd to create natural-language descriptions for analyst-written CrowdStrike Query Language (CQL) queries, enabling reverse labeling from existing structured artifacts.\footnote{\url{https://www.crowdstrike.com/en-us/blog/crowdstrike-journey-in-customizing-nvidia-nemotron-models/}} Before generation, the queries were deduplicated and sensitive values were scrubbed while preserving their structure. \ndd then generated and quality-filtered descriptions across analyst personas and complexity levels, and the accepted pairs were used to fine-tune Llama-3.3-Nemotron-Super-49B-v1.5. The resulting model achieved 96\% valid-query accuracy and a 2.50/5 semantic score, compared with 94\% and 2.35/5 for Claude Sonnet 4.5~\citep{crowdstrike2026}. This case shows how privacy-scrubbed enterprise records can be converted into validated supervised data for a specialized production task.

\FloatBarrier
\subsection{Search-Agent Trajectories}
\label{sec:app-search-agent}

\begin{figure}[!h]
\centering
\includegraphics[width=0.80\linewidth]{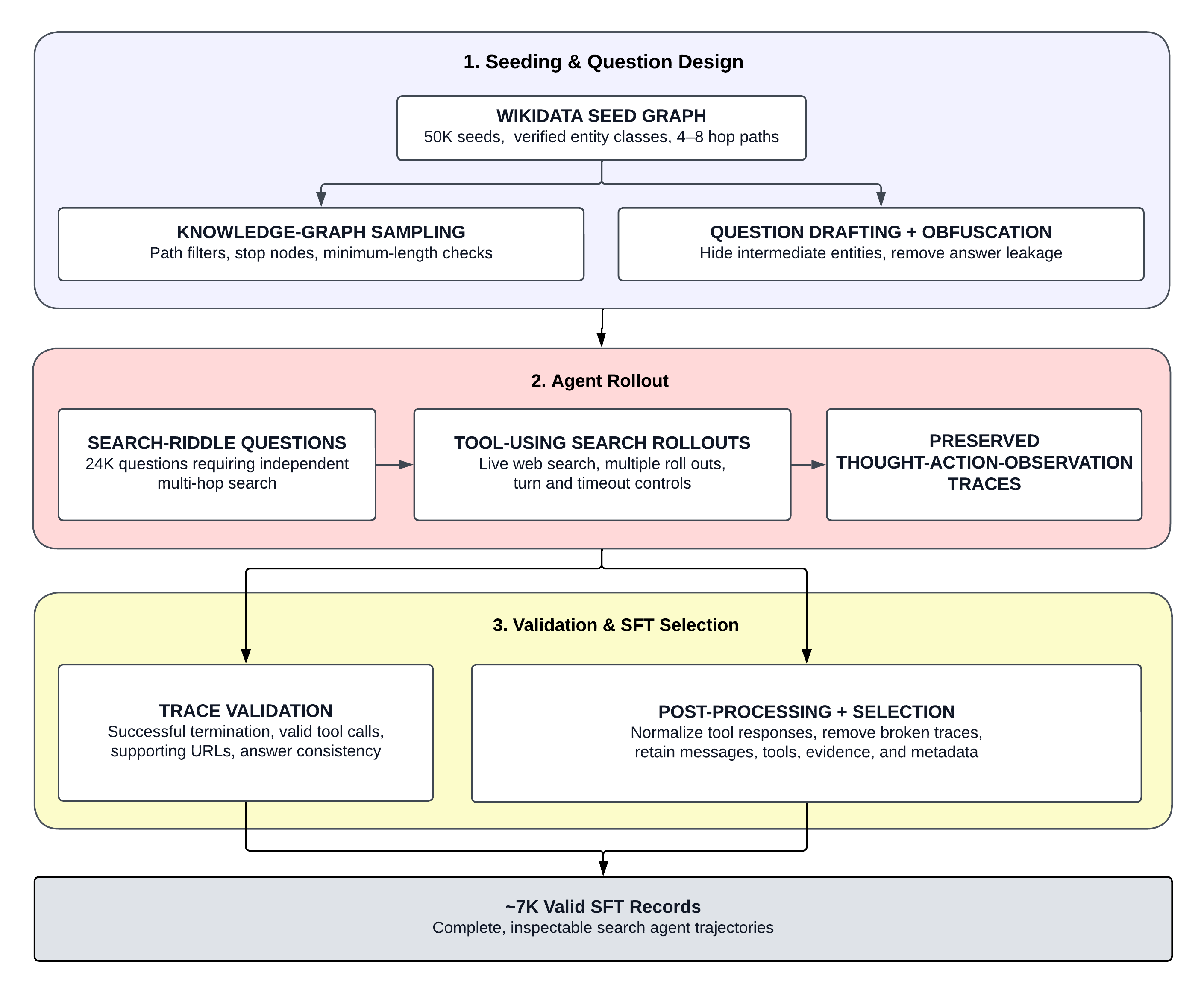}
\caption{\textbf{Search-agent synthetic-data pipeline.}
Wikidata paths become obfuscated search-riddle questions, tool-using rollouts
with preserved Thought--Action--Observation traces, and approximately 7K
validated SFT records.}
\label{fig:search-agent-workflow}
\end{figure}

\paragraph{Tool-trace preservation.}
Parametric knowledge is bounded by a model's training cutoff, while many user queries depend on current information or facts distributed across multiple sources. Answering these questions requires a search agent to issue queries, inspect results, refine its search strategy, and combine evidence over multiple hops. Figure~\ref{fig:search-agent-workflow} summarizes the progression from knowledge-graph seeds to validated, SFT-ready search trajectories.

Training this behavior requires complete interaction trajectories rather than final answers alone: each record must preserve the reasoning turns, search queries, tool calls, retrieved observations, supporting URLs, termination state, and synthesized response.

To generate this supervision, \ndd begins with four- to eight-hop paths through the Wikidata knowledge graph, converts each path into a natural-language question, and obfuscates intermediate entities so that the answer cannot be recovered by following explicit breadcrumbs. A rollout agent then attempts to solve the question using live web search through a Tavily MCP tool, while \ndd enforces tool allowlists, turn limits, and timeouts and captures the complete Thought--Action--Observation history. Successful trajectories are normalized into SFT-ready conversations that preserve the full search process. The developer note\footnote{\url{https://docs.nvidia.com/nemo/datadesigner/dev-notes/search-agent}} and search-agent recipe\footnote{\url{https://github.com/NVIDIA-NeMo/DataDesigner/blob/main/docs/assets/recipes/mcp_and_tooluse/search_agent.py}} provide implementations.

The pipeline converted 50K knowledge-graph seeds into 24K rollout questions and approximately 7K valid search trajectories, for an end-to-end yield of roughly 14\%. The accepted trajectories averaged approximately 12 tool calls, preserved the complete multi-turn search process, and were incorporated into Nemotron 3 Super supervised fine-tuning. This workflow demonstrates how \ndd can generate inspectable, long-horizon tool-use supervision at a scale that would be costly to annotate manually.

\FloatBarrier
\subsection{Long-Document Vision-Language Model Understanding}
\label{sec:app-long-document-vlm}

\paragraph{Iterative multimodal capability targeting.}
Long-document visual understanding requires models to interpret text, layout,
tables, charts, diagrams, and evidence distributed across many PDF pages.
The dataset must keep document and page identifiers, rendered page images,
optical character recognition (OCR) text where used, questions, answers,
reasoning traces, question-type labels, and judge decisions attached throughout
generation and filtering.

Rather than relying on one monolithic pipeline, the \ndd workflow evolved
through four complementary generation streams: OCR-grounded text QA,
classification-filtered visual QA, general single-page QA, and grouped-page
QA, with the last stream divided into multi-page and whole-document paths.
Evaluation failures determined which
document types and reasoning modes were added next, while an independent
vision-language model (VLM) judge filtered examples for correctness, question
quality, visual grounding, format compliance, and training-signal strength.
The developer note and nine-stage recipe suite document these patterns and
provide implementation examples.\footnote{\url{https://docs.nvidia.com/nemo/datadesigner/dev-notes/vlm-long-document-understanding}}\footnote{\url{https://github.com/NVIDIA-NeMo/DataDesigner/tree/main/docs/assets/recipes/vlm_long_doc}}

\begin{figure}[!htbp]
\centering
\includegraphics[width=0.94\linewidth]{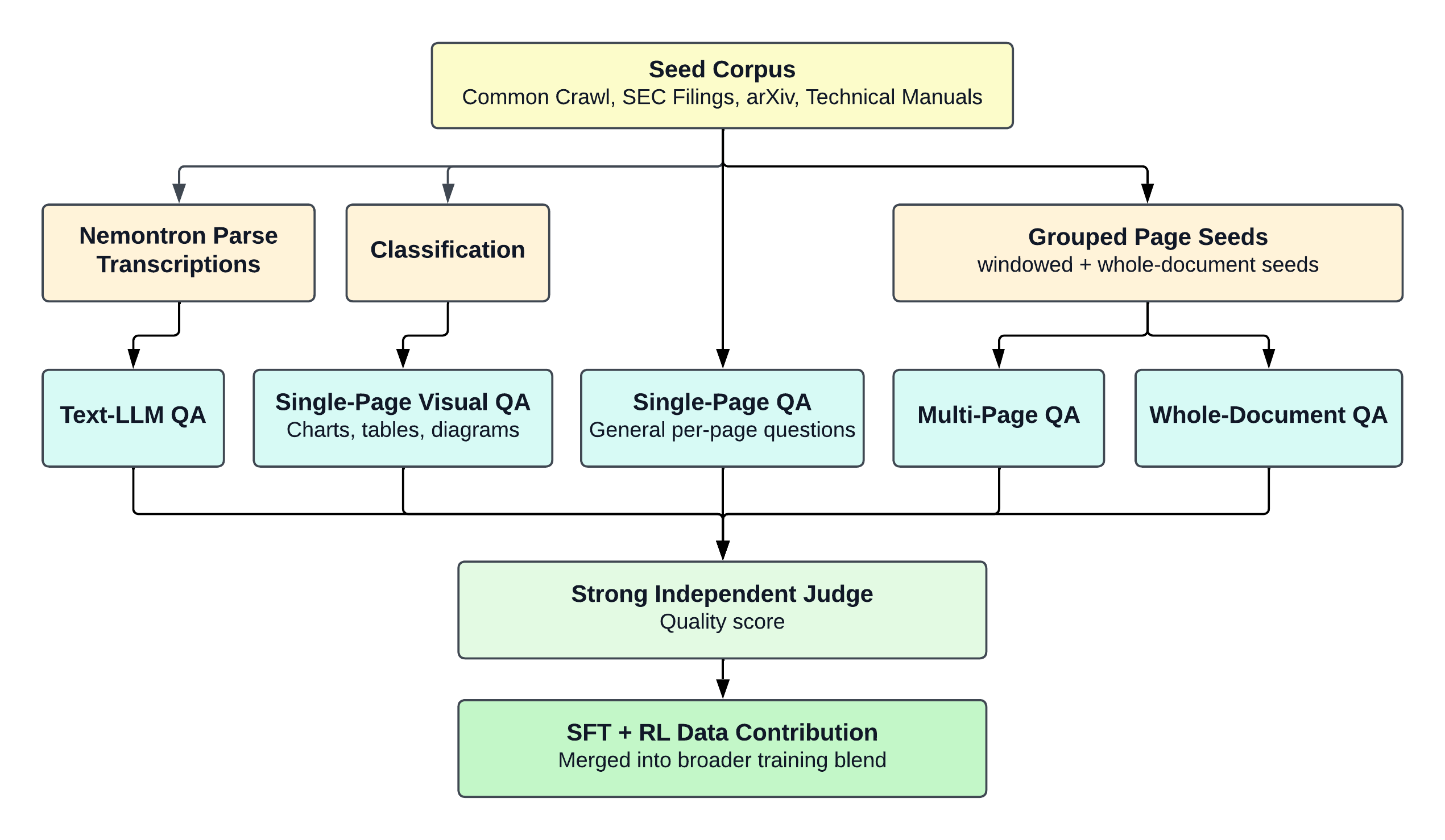}
\caption{\textbf{Long-document VLM synthetic-data pipeline.}
A shared seed corpus is transformed through Nemotron Parse transcription,
visual-content classification, and grouped-page preparation into complementary
text, single-page visual, single-page, multi-page, and whole-document QA paths.
A strong independent judge scores quality before selected records contribute
to the broader SFT and RL training blend.}
\label{fig:long-document-vlm-sdg-pipeline}
\end{figure}

Figure~\ref{fig:long-document-vlm-sdg-pipeline} summarizes how the shared
document corpus feeds complementary QA-generation paths and independent
quality scoring before selected records enter the broader SFT and RL training
blend.

The workflow produced approximately 11.4M visual question-answer pairs, representing about 45B tokens when questions, answers, reasoning traces, and vision tokens are included. Across successive data-generation, supervised fine-tuning, and reinforcement-learning interventions, experimental checkpoints improved from 26.32\% to 59.00\% on MMLongBench-Doc~\citep{ma2024mmlongbenchdoc}. The released Nemotron 3 Nano Omni model reports 57.5\% in reasoning-on mode~\citep{nvidia2026nemotron3nanoomni}.

The principal systems result is the iterative loop itself: evaluation exposed
missing capabilities, \ndd generated targeted supervision for them, and the
process repeated at increasing levels of document and reasoning complexity.

\FloatBarrier
\subsection{Prompt Robustness Through Diverse Instructions}
\label{sec:app-prompt-robustness}

\paragraph{Controlled linguistic diversity.}
Prompt-robustness data reduces sensitivity to superficial changes in instruction wording, tone, placement, and answer format while leaving the underlying task unchanged. The required artifact is not a new collection of problems, but a validated pool of interchangeable preambles and format instructions that can be applied to existing SFT and RL examples. Starting from 50 combinations of hand-written preamble anchors and regex-paired answer formats, \ndd samples six variation axes spanning sentence type, tone, strictness, verbosity, domain, and prompt ordering. Separate generation columns produce the preamble, format instruction, and assembled prompt, after which judges filter for format compliance, regex alignment, ordering coherence, and linguistic quality. The complete workflow is available in the prompt-sensitivity developer note.\footnote{\url{https://docs.nvidia.com/nemo/datadesigner/dev-notes/prompt-sensitivity}}

Early Nemotron checkpoints showed accuracy swings of as much as 15 percentage points across prompt phrasings for identical questions. The accepted prompt pool was incorporated into training mixtures using a controlled ratio of canonical and varied instructions; it used 25\% canonical and 75\% varied prompts. This recipe was used in Nemotron 3 Nano training, where we observed a roughly 2$\times$ reduction in prompt-sensitivity variation across GPQA, MMLU-Pro, competition mathematics, and LiveCodeBench~\citep{nvidia2025nemotron3nano}.

\FloatBarrier
\subsection{Nemotron-Personas}
\label{sec:app-personas}

\paragraph{Objective and pipeline.}
\begin{wrapfigure}{r}{0.50\textwidth}
\vspace{-0.7\baselineskip}
\centering
\includegraphics[width=\linewidth]{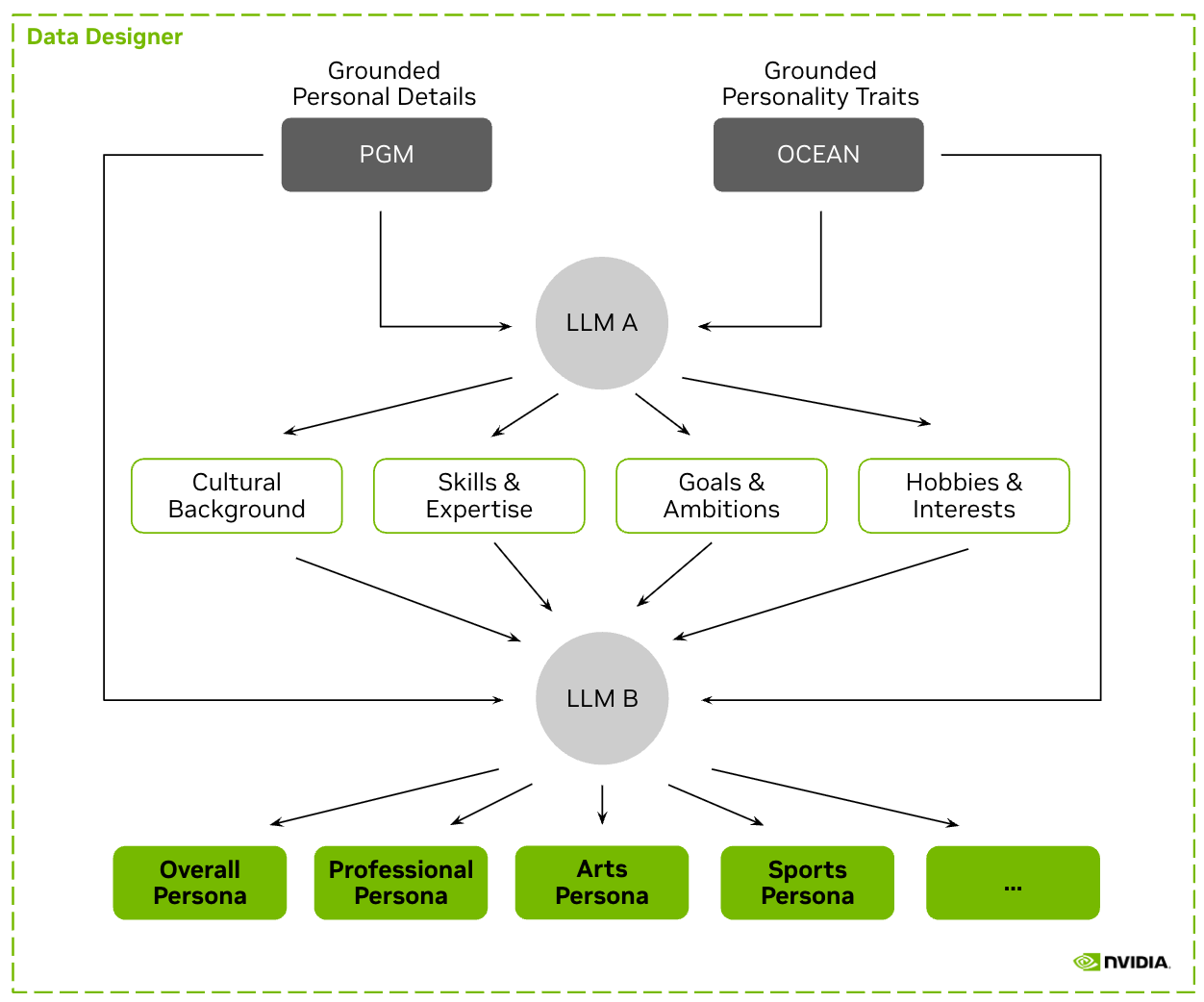}
\caption{\textbf{Grounded persona generation.}
A census-derived PGM and OCEAN descriptors feed two structured LLM stages.
Each stage may use an independent model alias.}
\label{fig:nemotron-personas-approach}
\vspace{-0.5\baselineskip}
\end{wrapfigure}
Nemotron-Personas introduces synthetic personas grounded in demographic,
geographic, and personality-trait statistics to capture the diversity and
complexity of a population. In contrast to earlier
persona-driven data synthesis~\citep{chan2024personahub}, which scales
perspective diversity through
web-derived personas, Nemotron-Personas grounds regional populations in census
and administrative statistics. This grounding preserves real-world structure
while diversifying synthetic data. It also reduces skew, enables targeted
coverage of lower-frequency groups, and can help mitigate bias and model
collapse~\citep{shumailov2024modelcollapse,dohmatob2024strong}. Regional
datasets also support multilingual and sovereign AI development.

Figure~\ref{fig:nemotron-personas-approach} summarizes the Data Designer
pipeline. For Nemotron-Personas-USA, the probabilistic graphical model (PGM)
encodes selected joint distributions from U.S. Census Bureau tables and
name-frequency data~\citep{nemotronpersonasusa2025}.
OCEAN descriptions add a separate psychometric input. One LLM call creates an
intermediate profile and a second produces context-specific personas.
\ndd resolves the column graph and enforces output schemas. The pipeline
follows the
compound AI framing~\citep{zaharia2024compoundai}.

\paragraph{Distributional quality and localization.}
Nemotron-Personas-USA contains one million records mirroring complex real-world distributions and patterns like non-Gaussian age profiles, life-stage changes in marital status, and
regional variations in education. Coverage includes approximately 29K ZIP Code
Tabulation Areas and more than 560 occupations. An LLM-only baseline failed to produce both simple distributions (e.g. age) and more complex joint disributions (e.g., educational attainment by geographic location and sex) ~\citep{nemotronpersonasusa2025}.

\begin{figure}[!htbp]
\centering
\includegraphics[width=0.82\linewidth]{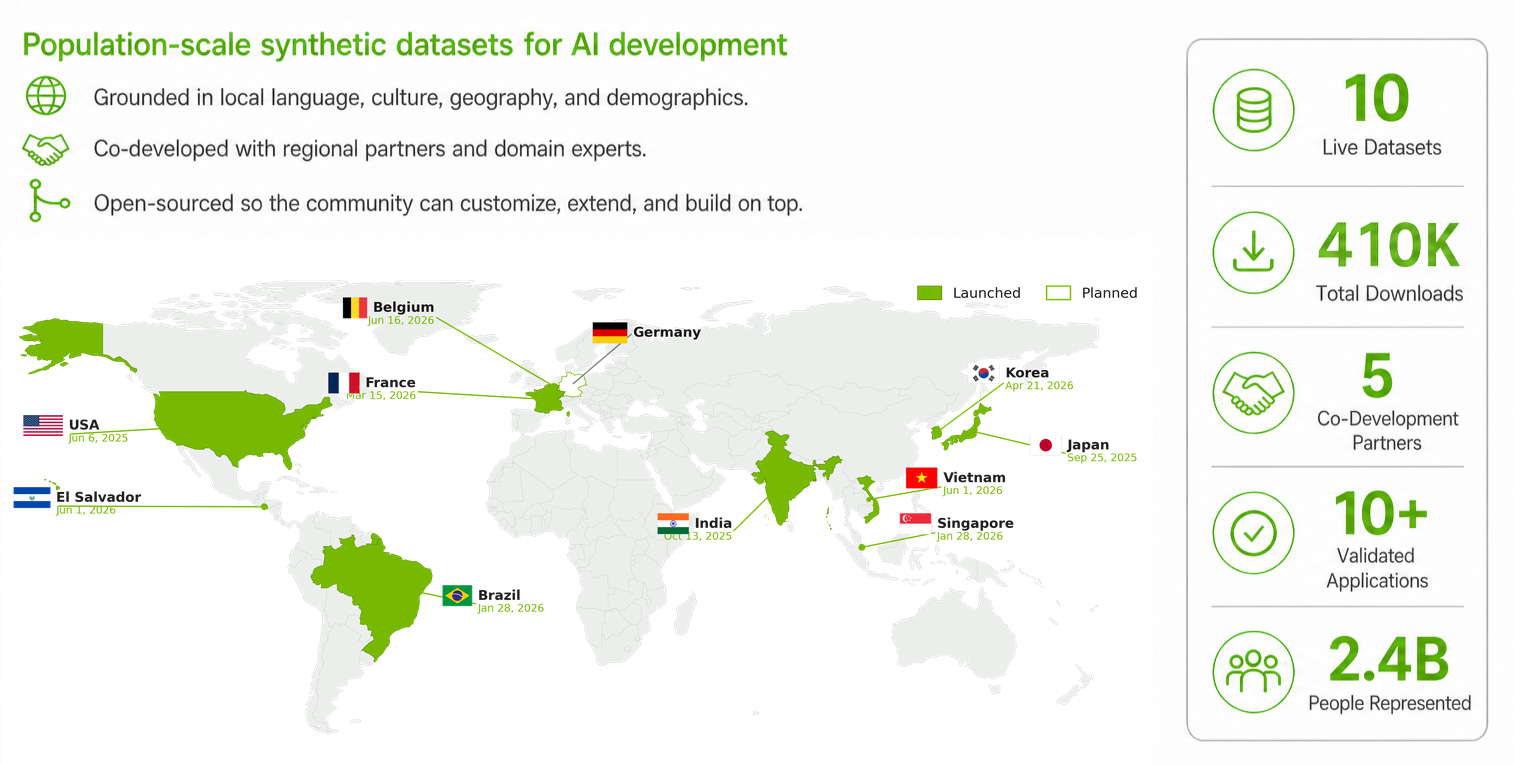}
\caption{\textbf{Nemotron-Personas global footprint as of August 12, 2026.}
Ten live datasets cover regions representing approximately 2.4B people. The
collection had 410K downloads, five co-development partners, and more than ten
validated applications.}
\label{fig:nemotron-personas-footprint}
\end{figure}

The public collection, summarized in
Figure~\ref{fig:nemotron-personas-footprint}, now contains ten regional
datasets and 15 language or script
editions~\citep{nemotronpersonascollection2025}.
India spans English and two Hindi scripts. Belgium provides Dutch, French,
German, and English editions. Locale-specific fields, regional review, and
language-specialist models preserve distinctions that translation alone cannot
supply~\citep{nemotronpersonasindia2025,nemotronpersonasbelgium2026}.

\paragraph{A building block for training and evaluation.}
Nemotron-Personas provides reusable grounding for both training-data synthesis
and evaluation. Sampling the persona before generation fixes the user
perspective while the downstream task changes.
Nemotron 3 Super uses Nemotron-Personas-USA for long-context aggregation and
simulated tool users, while personas from the broader collection diversify
formal-logic scenarios~\citep{nvidia2026nemotron3super}. Nemotron 3 Nano
extends the pattern to precise instruction following
\citep{nvidia2025nemotron3nano}. Its training recipe also uses
Nemotron-Personas-USA for general chat and safety data
\citep{nvidia2025nemotron3nanomodel}.

Each dataset in the quickly expanding collection is produced in one or more region-specific languages. This enables high-quality multilingual data and sovereign model development. For example, a 4B-token Japanese
tool-calling dataset grounded in Nemotron-Personas-Japan was used to further post-train
NVIDIA-Nemotron-Nano-9B-v2-Japanese, which ranked first among sub-10B models
on Nejumi Leaderboard 4~\citep{nvidia2026nemotronnano9bjapanese}.
Nemotron-Personas-Japan also enabled NTT DATA to expand 450 legal examples
into more than 138K training examples for tsuzumi 2. Synthetic-data SFT
increased QA accuracy from 15.3\% to 79.3\%
\citep{nttdata2026tsuzumi2personas}.
Ai2 used Nemotron-Personas-USA to generate approximately 220K verified
instruction records for each Olmo 3 Think SFT mixture
\citep{teamolmo2025olmo3}, and AMALIA generated European Portuguese
instruction-following and mathematics data~\citep{simplicio2026amalia}.

Nemotron-Personas also defines controlled test populations.
APTO used Japan personas for elderly-user red teaming, reducing attack success
from 6\% to 0\% across 100 attack prompts after safety fine-tuning
\citep{nvidia2026apto}. AMemGym uses a 100K-persona pool to construct
long-horizon memory evaluations
\citep{cheng2026amemgym}. ProactBench uses 50 profiles to create 198
dialogues with 624 proactive-behavior triggers
\citep{harfi2026proactbench}, while PICon samples seven regional datasets to
test persona consistency over 50-turn conversations~\citep{kim2026picon}.

\paragraph{Open pipeline and customization.}
In addition to datasets, we have open-sourced the full Nemotron-Personas pipeline.The developer note~\citep{meyer2026designingpersonas} explains the
design, and its companion notebook~\citep{nvidia2026personasnotebook} provides
a runnable recipe for NGC sampling, structured LLM stages, and domain
extension. The recipe is implemented with Data
Designer~\citep{nvidia2025datadesigner}, leaving models, schemas, and
validation policies configurable.

Furthermore, our open-source SDG-PGMs~\citep{nvidia2026sdgpgms} provides a
lightweight path to complete customization. The Python framework constructs
PGMs from aggregate public or proprietary distributions and integrates their
samples into Data Designer through \texttt{PGMGenerator}.
\texttt{PersonSamplerParams} remains the direct path for reusing released
persona assets.

Nemotron-Personas demonstrates this abstraction for synthetic people, but
SDG-PGMs is entity-agnostic. The same approach can model any entity with
structured relationships among its attributes. Examples include products and
transactions, as well as devices and proprietary business objects.

\FloatBarrier

\subsection{Broader applicability and SDG Design Principles}
The applications above are representative rather than exhaustive. Beyond these examples, \ndd has supported retrieval-data generation, scientific-reasoning datasets, and other domain-specific SDG workflows through task-specific seed sources, plugins, validators, processors, and export schemas. Across research and production workflows, NDD has supported the generation of approximately 10 trillion tokens of synthetic data, demonstrating that its programming model scales beyond individual recipes to sustained real-world use.

Taken together, these applications show that effective SDG begins by defining the dataset's intended coverage rather than maximizing generation volume. Samplers make the desired variation explicit, representative seeds anchor generation in factual or domain-specific context, and difficulty strata can turn a flat collection into a deliberate curriculum. The same principle applies to implementation: deterministic decisions should use samplers, expressions, or structured columns, reserving open-ended LLM generation for fields that require it.

Generation should then be treated as an iterative selection process. The text-to-SQL workflow illustrates the value of generating broadly and retaining only candidates that pass a quality waterfall, while the structured-output workflows show that objective requirements are best enforced with parsers, schemas, linters, and custom validators. LLM judges should be reserved for semantic criteria and, where possible, separated from the generator to reduce shared blind spots. Previewing a small sample, inspecting both accepted and rejected records, and revising the design before scaling completes the loop.

% AUTHOR TODO -- remove before camera-ready submission.
% 1. Verify that all technical reports and benchmarks use the canonical
%    BibTeX keys from the Overleaf project.
% 2. Confirm that the main include list uses the combined Applications file
%    and that the previous use-case and case-study files are no longer included.
% 3. Remove this TODO block before submission.

\section{Conclusion}
\label{sec:conclusion}

We presented \ndd, an open-source framework for designing, generating, and validating multimodal synthetic datasets. With \ndd, both humans and agents can define dataset columns, samplers, dependencies, and quality gates; inspect accepted and rejected records; and turn results into reusable artifacts. They can start with a small preview, revise the design based on what they observe, and then run it at scale without rewriting the workflow.

The case studies highlight several ways \ndd can be used, but they are not meant to be an exhaustive list. Its flexible and extensible programming model can be adapted across data types, modalities, validation strategies, and downstream uses. The workflows described here have contributed to Nemotron model development and enterprise applications. Their outcomes are necessarily measured differently, but together they support a common conclusion: synthetic-data generation should be treated as an explicit, inspectable, and reusable data-design process rather than a collection of isolated prompts.

This matters because the choices made in a synthetic-data pipeline influence what a model ultimately learns. Building the data pipeline is part of building the model. \ndd helps make that level of care practical for both humans and agents.

\section{Limitations}
\label{sec:limitations}

\paragraph{Real data and validation still set the boundary.} Synthetic data is most useful when there is enough domain knowledge, seed material, or validation logic to constrain generation. It is weaker when the target phenomenon is new, poorly understood, rapidly changing, or difficult to validate automatically. A generation pipeline can only optimize for the checks it contains: if a validator misses a semantic error, rewards superficial style, or encodes a biased rubric, the accepted dataset will inherit those weaknesses. This is sharpest for multimodal data, where visual grounding cannot be checked deterministically and a vision-language judge is itself an imperfectly calibrated instrument; OCR errors in document pipelines propagate into
questions and answers that remain well formed. Recursive use of untracked model outputs can also erode distributional coverage, while accumulation with real data and verification can mitigate that risk~\citep{shumailov2024modelcollapse,gerstgrasser2024modelcollapse,feng2024verification}. Real data and human review remain essential for discovering unknown failure modes, calibrating LLM judges in high-stakes settings, and measuring deployment behavior.

\paragraph{Filtering changes the distribution.} Rejection sampling improves average record quality, but it can also remove difficult, rare, or dialect-specific cases. A dataset that is well balanced before validation may become skewed after filtering, and samplers do not guarantee that a synthetic dataset matches a real population or task distribution. Workflows that depend on distributional fidelity should use representative seed data or external statistics, then audit accepted and rejected records rather than only the final dataset.

\paragraph{Endpoints, cost, and reproducibility are part of the experiment.}
Multi-stage pipelines often generate more candidates than they retain, which increases compute cost, wall-clock time, and operational complexity. The async runtime and adaptive concurrency layer reduce orchestration overhead, but they do not eliminate the cost of model calls or validation. Because endpoint and tool behavior can also change generated data, reproducibility should follow the auditable-recipe guidance in Section~\ref{sec:export-reproducibility}.

\paragraph{Privacy, leakage, and downstream utility require separate checks.} Synthetic data can reduce exposure to sensitive records, especially when generated from aggregate statistics or scrubbed seed data, but it should not automatically be treated as private. Pipelines that condition on real records can leak information if prompts, outputs, traces, or intermediate columns preserve sensitive details. Generated data can also overlap with benchmark-like sources, and a dataset can satisfy its schema while failing to improve a model. Privacy-sensitive workflows therefore need explicit scrubbing, access controls, access logging, review, benchmark-overlap checks, and task-specific downstream evaluation.

\section{Broader Impact}
\label{sec:broader-impact}

\ndd can reduce some manual annotation effort when seed data and validators are available. Reusable generation and validation workflows can also reduce the marginal cost of producing new dataset variants, although total savings depend on model-inference, validation, and human-review costs. This can help researchers, startups, and organizations with constrained annotation budgets build datasets for domains where expert labels are scarce. It can also support fairness and robustness work by making underrepresented cases easier to specify, generate, inspect, and test.

The same capabilities can be misused. Synthetic data pipelines can generate persuasive text, phishing-like examples, biased personas, or task-specific datasets for harmful automation. They can also create a false sense of validity if generated records are treated as real observations rather than artifacts produced under a specification. Responsible use requires provenance tracking, validation, bias assessment, privacy review, access controls for sensitive pipelines, and clear documentation of what the data is and is not intended to represent.

\section{Availability}
\label{sec:availability}

\ndd is available as open-source software under the Apache 2.0 license in its GitHub repository. This report describes version 0.9.0, and the library supports Python 3.10 and
later. Reference material, installation instructions, tutorials, and recipes are available in the \href{https://nvidia-nemo.github.io/DataDesigner/latest/}{project documentation}. The library can be used with user-provided model endpoints, including hosted APIs, centralized LLM gateways, and OpenAI-compatible self-hosted servers.

\section{Acknowledgments}
\label{sec:acknowledgments}

We thank the NVIDIA NeMo team and the broader Data Designer community for building and documenting the framework described in this report. We also thank the Nemotron teams whose implementations and reports provided the concrete workflows summarized here, including retrieval, text-to-SQL, structured-output, search-agent, deep-research, science-reasoning, long-document VLM, and persona-generation efforts. Finally, we thank the users and partner teams who used Data Designer to create and validate datasets across research and production workflows, helping exercise the framework at scale.

We are grateful to CrowdStrike for publishing details of the natural-language-to-CQL collaboration, which illustrates the role of synthetic data in a production security workflow. We also thank Perplexity AI for its collaboration on model-usability requirements and evaluations that informed the Nemotron 3 Ultra workflow described in Section~\ref{sec:app-structured-outputs}.

\begingroup
\sloppy
\bibliography{references}
\bibliographystyle{references}
\endgroup

\clearpage
\appendix

\section{Installation Examples}
\label{app:installation-examples}

This appendix collects the small setup and first-run material that is useful for readers who want to reproduce the programming model without interrupting the main systems narrative. \ndd is a Python package that calls user-provided model endpoints and provides default provider configuration for common hosted services.

\begin{lstlisting}[language=DDSh,caption={Installation and provider-configuration checks. API-key values are placeholders.},label={lst:appendix-install}]
pip install data-designer

export NVIDIA_API_KEY="your-api-key-here"
export OPENAI_API_KEY="your-openai-api-key-here"
export OPENROUTER_API_KEY="your-openrouter-api-key-here"

data-designer config list
\end{lstlisting}

The smallest useful workflow is to declare a builder, add columns, preview a few records, and then revise the design before a full \texttt{create()} run. Listing~\ref{lst:appendix-minimal-preview} shows a greeting example and is intentionally limited to the preview checkpoint; Appendix~\ref{app:declarative-config-example} compares the same builder pattern with a compact YAML declaration.

\begin{lstlisting}[language=Python,caption={Minimal preview workflow. The prompt references the sampled \texttt{language} field with Jinja syntax, so the runtime can infer the column dependency.},label={lst:appendix-minimal-preview}]
import data_designer.config as dd
from data_designer.interface import DataDesigner

data_designer = DataDesigner()
config_builder = dd.DataDesignerConfigBuilder()

config_builder.add_column(
    dd.SamplerColumnConfig(
        name="language",
        sampler_type=dd.SamplerType.CATEGORY,
        params=dd.CategorySamplerParams(
            values=["English", "Spanish", "French", "German", "Italian"],
        ),
    )
)

config_builder.add_column(
    dd.LLMTextColumnConfig(
        name="greeting",
        model_alias="nvidia-text",
        prompt="Write a casual and formal greeting in {{ language }}.",
    )
)

results = data_designer.preview(config_builder)
results.display_sample_record()
\end{lstlisting}

The setup should be read as an interface example rather than a complete experimental protocol. For the reproducibility record required for larger runs, see Section~\ref{sec:export-reproducibility}.

\section{Configuration Examples}
\label{app:config-examples}

The examples below show declarative and builder representations alongside short configuration excerpts for common \ndd patterns: sampler-controlled generation, structured outputs, multimodal context, image generation, validation columns, processor-based export views, and MCP trace capture. They are not intended to be a complete recipe for any one public case study. String placeholders such as \texttt{"my-model-alias"}, \texttt{"my-image-model-alias"}, \texttt{"my-multimodal-model-alias"}, and \texttt{"demo-mcp"} denote user-configured aliases rather than built-in API names. The appendix provides a stable destination for code and configuration details that would otherwise crowd the main body.

\subsection{Declarative Configuration and Builder Equivalence}
\label{app:declarative-config-example}

The same dataset design can be represented as a compact declarative schema or
constructed through the Python builder. Listings~\ref{lst:declarative-yaml}
and~\ref{lst:declarative-python} compare the two forms; runtime calls such as
\texttt{preview()} and \texttt{create()} remain outside the dataset
specification in both cases.

\begin{lstlisting}[
    language=DDYaml,
    caption={Declarative YAML dataset configuration.},
    label={lst:declarative-yaml}
]
columns:
  - name: audience
    column_type: sampler
    sampler_type: category
    params:
      values: [developer, analyst, student]

  - name: goal
    column_type: sampler
    sampler_type: category
    params:
      values: [definition, comparison]

  - name: answer
    column_type: llm-text
    model_alias: nvidia-text
    prompt: >
      Write a {{ goal }} answer for a
      {{ audience }} about dependency-aware
      synthetic data generation.

processors:
  - name: chat_view
    processor_type: schema_transform
    template:
      messages:
        - role: user
          content: "Explain {{ goal }}."
        - role: assistant
          content: "{{ answer }}"
\end{lstlisting}

\newpage

\begin{lstlisting}[
    language=Python,
    caption={Equivalent Python builder configuration.},
    label={lst:declarative-python}
]
import data_designer.config as dd

builder = dd.DataDesignerConfigBuilder()
builder.add_column(dd.SamplerColumnConfig(
    name="audience",
    sampler_type=dd.SamplerType.CATEGORY,
    params=dd.CategorySamplerParams(
        values=["developer", "analyst", "student"],
    ),
))
builder.add_column(dd.SamplerColumnConfig(
    name="goal",
    sampler_type=dd.SamplerType.CATEGORY,
    params=dd.CategorySamplerParams(
        values=["definition", "comparison"],
    ),
))
builder.add_column(dd.LLMTextColumnConfig(
    name="answer",
    model_alias="nvidia-text",
    prompt=(
        "Write a {{ goal }} answer for a "
        "{{ audience }} about dependency-aware "
        "synthetic data generation."
    ),
))
builder.add_processor(dd.SchemaTransformProcessorConfig(
    name="chat_view",
    template={"messages": [
        {"role": "user", "content": "Explain {{ goal }}."},
        {"role": "assistant", "content": "{{ answer }}"},
    ]},
))
\end{lstlisting}

\subsection{Model Providers and Aliases}

A \texttt{ModelProvider} defines endpoint connectivity, while
\texttt{ModelConfig} objects assign inference policies to logical aliases used
by columns. Listing~\ref{lst:appendix-model-provider-alias-config} configures two
aliases with different temperatures for the same underlying model; the column
definitions depend only on those aliases.

\begin{lstlisting}[
    language=Python,
    caption={Provider configuration and two model aliases with distinct inference policies.},
    label={lst:appendix-model-provider-alias-config}
]
import data_designer.config as dd
from data_designer.interface import DataDesigner

provider = dd.ModelProvider(
    name="enterprise-gateway",
    endpoint="https://llm.example.com/v1",
    api_key="LLM_API_KEY",
)

generator = dd.ModelConfig(
    alias="generator",
    model="organization/model",
    provider=provider.name,
    inference_parameters=dd.ChatCompletionInferenceParams(
        temperature=0.9,
        max_parallel_requests=32,
    ),
)

critic = dd.ModelConfig(
    alias="critic",
    model="organization/model",
    provider=provider.name,
    inference_parameters=dd.ChatCompletionInferenceParams(
        temperature=0.1,
        max_parallel_requests=32,
    ),
)

data_designer = DataDesigner(model_providers=[provider])
config_builder = dd.DataDesignerConfigBuilder(
    model_configs=[generator, critic]
)

config_builder.add_column(dd.LLMTextColumnConfig(
    name="answer",
    model_alias="generator",
    prompt="Explain dependency-aware SDG.",
))

config_builder.add_column(dd.LLMTextColumnConfig(
    name="critique",
    model_alias="critic",
    prompt="Critique: {{ answer }}",
))
\end{lstlisting}

\subsection{Structured Output}

\texttt{LLMStructuredColumnConfig} supports either Pydantic models or JSON schemas as the output contract. Listing~\ref{lst:appendix-structured-config} shows a compact Pydantic-based excerpt.

\begin{lstlisting}[language=Python,caption={Sampler and structured-output excerpt. The shown model alias is a user-configured placeholder.},label={lst:appendix-structured-config}]
from pydantic import BaseModel, Field

import data_designer.config as dd


class Product(BaseModel):
    name: str
    description: str
    price: float = Field(ge=10, le=1000)


config_builder.add_column(dd.SamplerColumnConfig(
    name="product_category",
    sampler_type=dd.SamplerType.CATEGORY,
    params=dd.CategorySamplerParams(
        values=["Electronics", "Clothing", "Home Office"],
    ),
))
config_builder.add_column(dd.LLMStructuredColumnConfig(
    name="product",
    prompt="Create a {{ product_category }} product.",
    output_format=Product,
    model_alias="my-model-alias",
))
\end{lstlisting}

\subsection{Multimodal Context and Image Generation}

\texttt{ImageColumnConfig} generates an image through a configured image-model alias, while \texttt{multi\_modal\_context} attaches media from existing columns to a model request. Listing~\ref{lst:appendix-multimodal-image-config} combines the two patterns: sampler values control an image prompt, and a downstream vision-language column receives the generated image through \texttt{ImageContext}. The context declaration also creates a dependency on \texttt{scene\_image}, so caption generation waits for the image column without requiring imperative orchestration. The selected aliases must refer to endpoints that support image generation and image input, respectively.

\begin{lstlisting}[language=Python,caption={Image generation followed by multimodal text generation. Both aliases are user-configured placeholders.},label={lst:appendix-multimodal-image-config}]
builder.add_column(dd.SamplerColumnConfig(
    name="scene_type",
    sampler_type=dd.SamplerType.CATEGORY,
    params=dd.CategorySamplerParams(values=[
        "a warehouse loading area",
        "an urban intersection",
        "a mountain trail",
    ]),
))

builder.add_column(dd.ImageColumnConfig(
    name="scene_image",
    prompt=(
        "Create a photorealistic image of {{ scene_type }} "
        "with clear foreground and background details."
    ),
    model_alias="my-image-model-alias",
))

builder.add_column(dd.LLMTextColumnConfig(
    name="scene_caption",
    prompt=(
        "Write concise, accessible alt text for the supplied "
        "image. Mention the setting and the most important objects."
    ),
    model_alias="my-multimodal-model-alias",
    multi_modal_context=[
        dd.ImageContext(column_name="scene_image"),
    ],
))
\end{lstlisting}

For seed datasets that already contain image URLs, the same pattern can use
\nolinkurl{ImageContext(column_name="source_image", data_type=dd.ModalityDataType.URL)}.
Generated image paths are resolved when used as downstream context; externally
hosted URLs must remain accessible to the selected model endpoint.

\subsection{Validation Columns}

Validation columns attach executable quality checks to generated fields. Listing~\ref{lst:appendix-validation-config} uses the Python-code validator pattern; the same validator interface also supports SQL dialect validation, local callables, and remote HTTP validators.

\begin{lstlisting}[language=Python,caption={Code-generation and validation-column excerpt. The validation result is retained because \texttt{drop=False}.},label={lst:appendix-validation-config}]
builder.add_column(
    dd.LLMCodeColumnConfig(
        name="sorting_algorithm",
        prompt="Write a Python function to sort a list using bubble sort.",
        code_lang="python",
        model_alias="my-model-alias",
    )
)

builder.add_column(
    dd.ValidationColumnConfig(
        name="code_validation",
        target_columns=["sorting_algorithm"],
        validator_type="code",
        validator_params=dd.CodeValidatorParams(
            code_lang=dd.CodeLang.PYTHON,
        ),
        batch_size=10,
        drop=False,
    )
)
\end{lstlisting}

\subsection{Export Views}

Processors transform generated batches or final datasets outside the per-column generation path. The built-in drop-columns processor removes intermediate fields from the primary output, while the schema-transform processor writes an additional transformed dataset view. Listing~\ref{lst:appendix-processor-config} shows both patterns.

\begin{lstlisting}[language=Python,caption={Processor excerpt for cleanup and a chat-format export view.},label={lst:appendix-processor-config}]
builder.add_processor(dd.DropColumnsProcessorConfig(
    name="cleanup",
    column_names=["scratch_work", "raw_context"],
))

builder.add_processor(dd.SchemaTransformProcessorConfig(
    name="chat_format",
    template={"messages": [
        {"role": "user", "content": "{{ question }}"},
        {"role": "assistant", "content": "{{ answer }}"},
    ]},
))
\end{lstlisting}

\subsection{MCP Tool Traces}

For tool-use data, \ndd can expose Model Context Protocol tools to an LLM column and capture the conversation as a side-effect trace column. Listing~\ref{lst:appendix-mcp-trace-config} is an excerpt of the basic MCP pattern: the provider name must correspond to a local or remote MCP provider registered by the application before execution, and the resulting trace is written to \texttt{fact\_response\_\_trace}.

\begin{lstlisting}[language=Python,caption={MCP tool-use and trace-capture excerpt. The \texttt{with\_trace} setting requests full message history.},label={lst:appendix-mcp-trace-config}]
tool_config = dd.ToolConfig(
    tool_alias="basic-tools",
    providers=["demo-mcp"],
    allow_tools=["get_fact"],
    max_tool_call_turns=5,
    timeout_sec=30.0,
)

builder = dd.DataDesignerConfigBuilder(tool_configs=[tool_config])

builder.add_column(
    dd.LLMTextColumnConfig(
        name="fact_response",
        model_alias="my-model-alias",
        prompt=(
            "Use the get_fact tool to look up information about "
            "'{{ topic }}', then write a one-sentence summary."
        ),
        system_prompt=(
            "You must call the get_fact tool before answering. "
            "Only use information from tool results."
        ),
        tool_alias="basic-tools",
        with_trace=dd.TraceType.ALL_MESSAGES,
    )
)
\end{lstlisting}

\subsection{Minimal Column-Generator Plugin}
\label{app:plugin-example}

A reusable column type consists of a configuration class, an implementation,
a plugin descriptor, and a package entry point. Listing~\ref{lst:appendix-plugin-implementation}
shows these pieces for a small plugin that multiplies each row index by a
configurable value, followed by its use in a dataset configuration. The files
are shown together for brevity but remain separate in the installed package.

\begin{lstlisting}[
    language=Python,
    commentstyle=\color{ddcodecomment},
    caption={Core components of a minimal column-generator plugin and its use in a dataset configuration.},
    label={lst:appendix-plugin-implementation}
]
# config.py
from typing import Literal
from data_designer.config.base import SingleColumnConfig

class IndexMultiplierColumnConfig(SingleColumnConfig):
    column_type: Literal["index-multiplier"] = "index-multiplier"
    multiplier: int = 2

    @property
    def required_columns(self) -> list[str]:
        return []

    @property
    def side_effect_columns(self) -> list[str]:
        return []

# impl.py
import pandas as pd
from data_designer.engine.column_generators.generators.base import (
    ColumnGeneratorFullColumn,
)

from data_designer_index_multiplier.config import (
    IndexMultiplierColumnConfig,
)

class IndexMultiplierColumnGenerator(
    ColumnGeneratorFullColumn[IndexMultiplierColumnConfig]
):
    def generate(self, data: pd.DataFrame) -> pd.DataFrame:
        data[self.config.name] = data.index * self.config.multiplier
        return data

# plugin.py
from data_designer.plugins import Plugin, PluginType

plugin = Plugin(
    config_qualified_name=(
        "data_designer_index_multiplier.config."
        "IndexMultiplierColumnConfig"
    ),
    impl_qualified_name=(
        "data_designer_index_multiplier.impl."
        "IndexMultiplierColumnGenerator"
    ),
    plugin_type=PluginType.COLUMN_GENERATOR,
)

# pyproject.toml
# [project.entry-points."data_designer.plugins"]
# index-multiplier = "data_designer_index_multiplier.plugin:plugin"

# Dataset configuration
from data_designer_index_multiplier.config import (
    IndexMultiplierColumnConfig,
)

builder.add_column(IndexMultiplierColumnConfig(
    name="scaled_index",
    multiplier=5,
))
\end{lstlisting}

\clearpage
\section{Model Runtime and Request Admission}
\label{app:model-runtime}

This appendix expands the model-runtime summary in Section~\ref{sec:models-and-providers}. These mechanics determine how a ready model-backed cell reaches an external endpoint; they do not change the declarative column graph or its data-design semantics.

\subsection{Model Facade and Client Stack}
\label{sec:model-facade-client-stack}

\begin{figure}[!ht]
\centering
\includegraphics[width=0.47\linewidth]{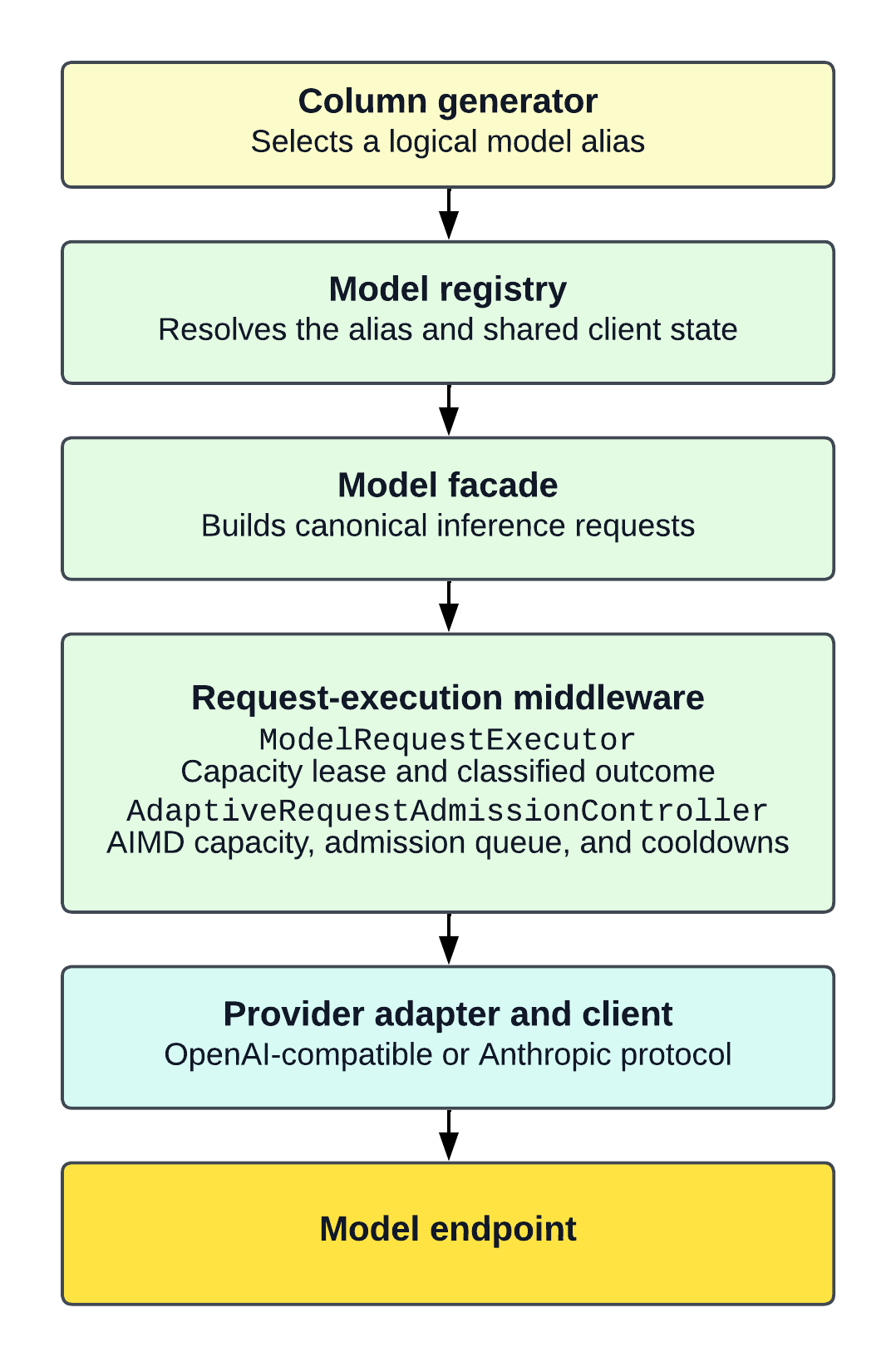}
\caption{\textbf{Model client stack.}
A column generator resolves a logical model alias through the registry
and facade. The request executor acquires capacity from the shared
adaptive admission controller before invoking the provider client, then
releases the lease with the classified outcome. AIMD feedback, admission
queues, and cooldowns regulate concurrency at the endpoint boundary.}
\label{fig:model-client-stack}
\end{figure}

At runtime, the \texttt{ModelRegistry} indexes model configurations by alias and lazily constructs a \texttt{ModelFacade} when an alias is first requested. The registry provides a single resolution point for column generators and coordinates shared request-capacity state for aliases that target the same provider and model. Column generators consequently request a model by alias rather than selecting an HTTP adapter or managing a client lifecycle directly. Figure~\ref{fig:model-client-stack} summarizes this stack and its request-admission path.

The \texttt{ModelFacade} is the common inference interface exposed to the engine. It provides blocking and awaitable operations for chat completion, embeddings, and image generation, and combines model-level inference parameters with provider-wide request extensions. It centralizes prompt and message construction, output parsing, structured-response validation, correction messages, conversation restarts, multimodal context, MCP tool-use loops, trace construction, and model-usage accounting.

For tool-enabled generation, the facade supplies tool schemas to the model, executes requested tools through the MCP integration, returns observations to the conversation, and continues until the model produces a final response or reaches the configured tool-turn limit. For structured generation, a parsing failure may trigger an in-conversation correction or a complete conversation restart according to \texttt{RunConfig}. The resulting message history can be retained as a trace column rather than discarded after the final value is produced.

Below the facade, provider adapters translate canonical \ndd request and response types into the provider's wire format. Native clients support OpenAI-compatible APIs and the Anthropic Messages API. Hosted services, centralized gateways, and self-hosted OpenAI-compatible servers therefore share the same registry and facade path, while provider-specific response objects and error formats are normalized before returning to the engine.

\FloatBarrier
\subsection{Adaptive Request Admission and AIMD}
\label{sec:adaptive-request-admission}

\ndd inserts a request-admission layer between the facade and each provider
client. Its \texttt{ModelRequestExecutor} wraps the model-client interface and
intercepts each concrete outbound attempt. It maps the call to a canonical
request resource identified by provider, underlying model identifier, and
request domain---chat, embedding, image, or health check---and acquires a lease
from the shared \texttt{AdaptiveRequestAdmissionController} before invoking the
provider.

Each lease represents one admitted request and is released with a classified outcome. Success, rate limiting, timeout, provider failure, local cancellation, and unexpected failure all release the exact lease that was acquired. Requests that cannot immediately obtain capacity wait in an admission queue. This acquire--call--release boundary prevents permits from leaking on exceptions or cancellation and exposes in-flight calls, queued requests, cooldowns, and active capacity for observation.

The configured \texttt{max\_parallel\_requests} value supplies a hard concurrency ceiling for a provider--model pair. Below that ceiling, the controller uses additive-increase/multiplicative-decrease (AIMD) feedback to adapt to observed endpoint capacity. An optional startup ramp begins a request domain at one concurrent call and increases toward its configured limit. When the provider returns HTTP~429, the controller applies the provider's \texttt{Retry-After} interval, or a configured fallback cooldown, and multiplicatively reduces the active limit. After a configurable window of successful requests, it additively restores capacity until it reaches the ceiling.

Rate limiting is separated from ordinary retry behavior. HTTP~429 responses remain visible because they drive AIMD. Eligible transient connection failures and selected server errors follow a separate retry and exponential-backoff policy, with every concrete retry attempt passing through request admission. This prevents retries from creating additional load while an endpoint is already signaling that it is over capacity.

Aliases resolving to the same provider and underlying model share a global request cap, so separate generator and judge aliases cannot independently exceed the capacity assigned to the same upstream model. Adaptive state remains separate by request domain, allowing chat, embedding, image, and health-check traffic to respond to their own outcomes while staying bounded by the shared model-level limit. The dataset scheduler decides which cells are ready; request admission decides when a ready cell may issue a concrete provider request.

\clearpage
\section{Operational Deployment and Capacity Planning}
\label{app:operational-deployment-and-capacity-planning}

This appendix describes user-managed deployment patterns for running multiple
independent \ndd execution units. Scheduler configuration, resource allocation,
work partitioning, job submission, and output consolidation remain outside
\ndd and are supplied by the user. Within each process, \ndd compiles column
dependencies, schedules ready work, adapts request admission, applies retries
and validation, and materializes checkpoints and output artifacts.

A separate deployment plan determines how work is partitioned, how many jobs
are launched, which resources are assigned to each job, and how their outputs
are named and consolidated. This allows the same dataset configuration to run
in a workstation process, a batch allocation, or a set of scheduler-managed
jobs without embedding infrastructure-specific orchestration in the dataset
specification. Figure~\ref{fig:deployment-scale-out-patterns} summarizes the
single-job and sharded architectures introduced in
Section~\ref{sec:execution-and-deployment-boundary}. This appendix develops
their operational implications, including capacity measurement, scheduler
integration, artifact isolation, and reproducibility.

When a deployment plan launches multiple jobs, each job owns a separate \ndd
runtime, checkpoint state, and artifact namespace. The plan must assign
non-overlapping shards and distinct dataset names. Adaptive request-admission
state remains local to each process, so the deployment plan must also bound the
aggregate request ceilings of jobs that share a gateway or serving pool.

\subsection{Pilot-Based Capacity Planning}
\label{app:pilot-capacity-planning}

Efficient large-scale generation requires planning before scheduler submission. For a fixed \ndd configuration, operators should first run a few representative single-job pilots using the intended model providers, validation stages, trace settings, and artifact format.  The pilots should record at least:

\begin{itemize}
  \item accepted records per unit time and model calls per accepted record;
  \item request latency, retry counts, throttling responses, and cooldown time;
  \item host memory utilization and local or remote I/O volume, together with CPU utilization when local validators, processors, or serialization are compute-intensive;
  \item for self-hosted model endpoints, per-GPU utilization and GPU memory usage;
  \item checkpoint size, artifact growth, and the fraction of rejected records.
\end{itemize}

For self-hosted endpoints, the pilot matrix should also vary the serving configuration.  Relevant choices include tensor-parallel degree, the number of model replicas, server-side batching or token limits, and endpoint request concurrency.  Increasing tensor parallelism may be necessary to fit a model or reduce per-request latency, but cross-GPU communication means that it does not necessarily maximize aggregate throughput.  When the model fits on fewer GPUs, additional replicas may instead provide higher throughput.  Operators should select the configuration that provides the best sustained accepted-record throughput while maintaining acceptable latency, GPU memory headroom, and stability under the expected request mix.  These are serving-layer decisions; \ndd observes the resulting endpoint behavior but does not configure the model server's parallelism strategy.

The pilot measurements inform the job count and shard size selected by the deployment plan.  Throughput should not be assumed to scale linearly with the number of jobs, particularly when jobs share model endpoints, gateways, or storage.  The plan must account for aggregate endpoint throughput, provider quotas, available scheduler resources, storage and network bandwidth, and cost limits.  It should record the configuration version, shard size, per-job resources, endpoint topology, request ceilings, and artifact layout so that the generation can be reproduced or revised.

\subsection{Scheduler Integration}
\label{app:scheduler-integration}

The deployment plan may be implemented by any environment capable of launching isolated processes with shard-specific arguments.  For example, a Slurm job array can map one array task to one dataset shard:

\begin{center}
\begin{minipage}{0.96\linewidth}
\begin{verbatim}
#!/bin/bash
#SBATCH --job-name=dd-generate
#SBATCH --array=0-31
#SBATCH --cpus-per-task=16
#SBATCH --mem=64G

DATASET_ROOT="/some/path/my_dataset"

srun python run_shard.py \
  --shard-id "${SLURM_ARRAY_TASK_ID}" \
  --num-shards "${SLURM_ARRAY_TASK_COUNT}" \
  --artifact-path "${DATASET_ROOT}" \
  --dataset-name "${SLURM_ARRAY_TASK_ID}" \
  --resume if_possible
\end{verbatim}
\end{minipage}
\end{center}

Here, \texttt{run\_shard.py} is a user-supplied launcher; the command-line
arguments shown above belong to that launcher rather than to the \ndd CLI. The
launcher selects the assigned partition, constructs the shared \ndd
configuration, and invokes \texttt{create()} with the common root as
\texttt{artifact\_path} and the array task ID as \texttt{dataset\_name}. The API
constructs the base dataset path as
\texttt{artifact\_path/dataset\_name}; task 7 therefore writes beneath
\texttt{/some/path/my\_dataset/7}. Passing
\texttt{ResumeMode.IF\_POSSIBLE} or \texttt{ResumeMode.ALWAYS} allows a retry to reuse that shard's checkpoint state.

Because every task uses the same artifact root, the shard directories remain
grouped beneath one dataset namespace while the task-ID dataset names prevent
jobs from overwriting one another. Resource requests, array width, retry policy,
and dependency handling remain deployment decisions owned by the user and the
scheduler. A reliable sharded run also uses an immutable, identifiable
configuration, assigns non-overlapping shards, writes each job to an independent
artifact namespace, and retains shard metadata, rejection diagnostics, and
trace provenance during consolidation.

\end{document}